\PassOptionsToPackage{utf8}{inputenc}
\documentclass{bioinfo}
\copyrightyear{2015} \pubyear{2015}

\usepackage{pifont}
\usepackage{booktabs}       
\usepackage{url}            
\usepackage{amsfonts}       
\usepackage{nicefrac}       
\usepackage{amsmath}
\usepackage{graphicx}
\usepackage{bm, upgreek}
\usepackage{amssymb}
\usepackage{array}
\usepackage{multirow}
\usepackage{mathtools}
\usepackage{arydshln}
\usepackage{tabularx}
\usepackage{color}
\usepackage{soul}
\usepackage{caption}
\usepackage{xcolor}
\usepackage{comment}
\usepackage{adjustbox}
\usepackage{hyperref}
\usepackage{cleveref}
\usepackage[ruled,vlined]{algorithm2e}
\usepackage{makecell}
\usepackage{diagbox}
\usepackage{subfig}
\usepackage{xcolor, soul}

\usepackage{subcaption}

\definecolor{bluecolor}{HTML}{0000FF}
\definecolor{greencolor}{HTML}{8CD0A4}
\definecolor{yellowcolor}{HTML}{F9D17C}
\definecolor{redcolor}{HTML}{000000} 
\definecolor{exredcolor}{HTML}{FF0000}

\access{Advance Access Publication Date: Day Month Year}
\appnotes{Manuscript Category}

\begin{document}
\firstpage{1}
\subtitle{Data and text mining}
\title{Improving Medical Reasoning through Retrieval and Self-Reflection with Retrieval-Augmented Large Language Models}
\author[Jeong \textit{et~al}.]{Minbyul Jeong$^{\text{\sfb 1}}$, Jiwoong Sohn$^{\text{\sfb 1}}$, Mujeen Sung$^{\text{\sfb 2,}*}$, and Jaewoo Kang\,$^{\text{\sfb 1,3,}*}$}
\address{
$^{\text{\sf 1}}$Department of Computer Science, Korea University, Seoul 02841, Republic of Korea \\
$^{\text{\sf 2}}$School of Computing, Kyung Hee University, Gyeonggi-do 17104, Republic of Korea \\
$^{\text{\sf 3}}$AIGEN Sciences, Seoul 04778, Republic of Korea \\
}

\corresp{$^\ast$To whom correspondence should be addressed.}

\history{accepted on ISMB 2024}

\editor{Associate Editor: XXXXXXX}
\abstract{
Recent proprietary large language models (LLMs), such as GPT-4, have achieved a milestone in tackling diverse challenges in the biomedical domain, ranging from multiple-choice questions to long-form generations.
To address challenges that still cannot be handled with the encoded knowledge of LLMs, various retrieval-augmented generation (RAG) methods have been developed by searching documents from the knowledge corpus and appending them unconditionally or selectively to the input of LLMs for generation.
However, when applying existing methods to different domain-specific problems, poor generalization becomes apparent, leading to fetching incorrect documents or making inaccurate judgments. 
In this paper, we introduce \textbf{Self-BioRAG}, a framework reliable for biomedical text that specializes in generating explanations, retrieving domain-specific documents, and self-reflecting generated responses.
We utilize 84k filtered biomedical instruction sets to train Self-BioRAG that can assess its generated explanations with customized reflective tokens.
Our work proves that domain-specific components, such as a retriever, domain-related document corpus, and instruction sets are necessary for adhering to domain-related instructions.
Using three major medical question-answering benchmark datasets, experimental results of Self-BioRAG demonstrate significant performance gains by achieving a 7.2\% absolute improvement on average over the state-of-the-art open-foundation model with a parameter size of 7B or less.
\textcolor{redcolor}{
Similarly, Self-BioRAG outperforms RAG by 8\% Rouge-1 score in generating more proficient answers on two long-form question-answering benchmarks on average.
}
Overall, we analyze that Self-BioRAG finds the clues in the question, retrieves relevant documents if needed, and understands how to answer with information from retrieved documents and encoded knowledge as a medical expert does.
We release our data and code for training our framework components and model weights (7B and 13B) to enhance capabilities in biomedical and clinical domains.
\\
\textbf{Availability:} Self-BioRAG is available at \href{https://github.com/dmis-lab/self-biorag}{https://github.com/dmis-lab/self-biorag}\\
\textbf{Contact:} \href{mujeensung@khu.ac.kr}{mujeensung@khu.ac.kr}, \href{kangj@korea.ac.kr}{kangj@korea.ac.kr}\\
\textbf{Supplementary information:} Supplementary data are available at \textit{Bioinformatics}
online.
}

\maketitle
\section{Introduction}
\begin{figure*}[t]
  \centering
  \includegraphics[width=0.9\textwidth]{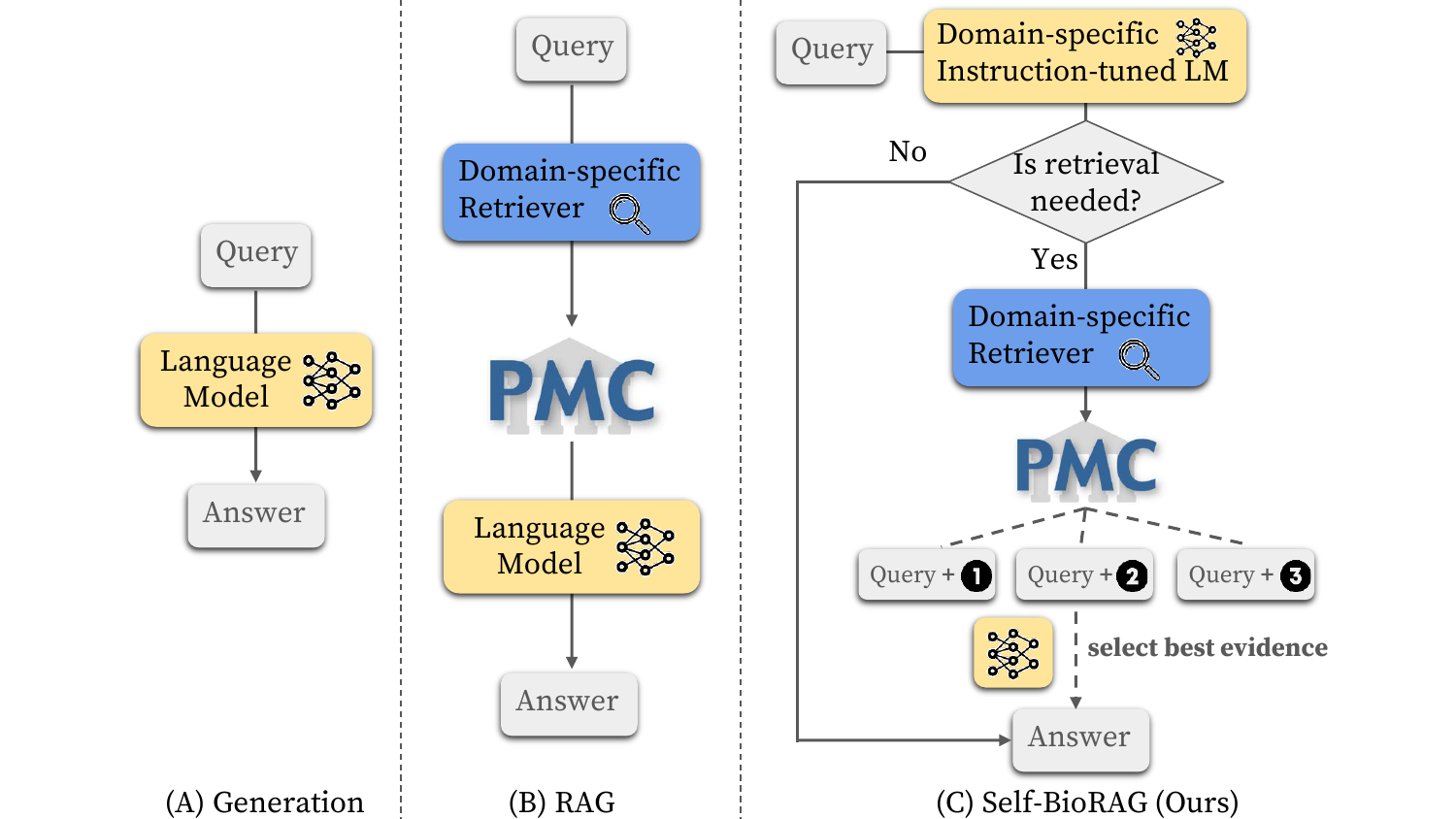}
  \caption{
  Comparison between three frameworks: generation using language model (LM), retrieval-augmented generation (RAG) using LM, and our Self-BioRAG.
  (A) depicts the process of sequence-to-sequence generation of LM. (B) The RAG framework first finds relevant documents from large-scale corpus such as PubMed Central and then provides the answer based on this factual content to address the shortage of scarce knowledge. (C) Initially, our domain-specific instruction-tuned model predicts whether retrieval is necessary. If a query doesn't require any retrieval of knowledge (factual content), it directly predicts the answer. However, if the query necessitates retrieval knowledge, Self-BioRAG utilizes the domain-specific retriever (MedCPT, in our case) to retrieve relevant documents.  After retrieving the top-$k$ evidence, the model selects the most pertinent evidence for the query. Ultimately, our language model is employed to select the best evidence and generate the answer based on the selected evidence and encoded knowledge.
  }\label{fig:intro_figure}
\vspace{-0.6cm}
\end{figure*}

The recent proprietary large language models (LLMs) such as ChatGPT~\citep{openai2023a}, GPT-4~\citep{openai2023b}, and BARD~\citep{bard2023} have succeeded in reaching near or comparable levels to human experts in solving many challenging problems, ranging from multi-choice question answering to long-form text generations. 
While these models exhibit high efficiency and demonstrate their versatility in various domains, they fall short in comprehensively covering user-dependent information such as patient reports with encoded knowledge.
These limitations can result in a groundless statement and inadvertent generation of false information, commonly known as the hallucination issue~\citep{wei2022chain, singhal2022large, cao-etal-2022-hallucinated, ji2023survey}.
To address this challenge, retrieval-augmented generation (RAG) enhances explainability for readers by supplying supporting facts that underpin the responses generated by LLMs~\citep{lewis2020retrieval, guu2020retrieval}.
As illustrated in Figure~\ref{fig:intro_figure}, various RAG frameworks search documents from the knowledge corpus such as Wikipedia and appending them unconditionally or selectively to the input of LLMs for generation.
\textcolor{redcolor}{
In alignment with this approach, the authors of \cite{asai2023self} introduce Self-RAG, which uses reflective tokens that learn to reflect on its generation process given a task input possessing the following capabilities: deciding when to use on-demand retrieval, assessing whether retrieved evidence provides useful information to solve the question, criticizing whether the evidence supports the answer, and judging whether the answer is a useful response to the question. }
However, using Self-RAG is unsuitable for domain-specific questions like biomedical or clinical domains which shows poor generalization, leading to fetching incorrect documents or making inaccurate judgments.

In this paper, we introduce Self-BioRAG, trained with a focus on biomedical and clinical text instructions, enabling it to address corresponding instructions adeptly.
It preserves generation quality and reasoning ability while incorporating on-demand retrieval and self-reflection capabilities.
Note that we use the term \textit{reasoning} to indicate that Self-BioRAG can provide explanations on answers.
To build a Self-BioRAG framework, four essential components are required: (1) biomedical instruction sets, (2) biomedical retriever, (3) self-reflection language model, and (4) domain-specific instruction-tuned language model.
We initially construct instruction sets focused on biomedical and clinical text.
In addition to the distributed MoL-instructions~\citep{fang2023mol} and MedInstruct~\citep{zhang2023alpacare}, we synthetically generate an additional 18k biomedical and clinical instructions following the Self-Instruct~\citep{wang2022self}.
By combining three datasets, we could construct 120k instruction sets addressing various biomedical instructions, including information extraction, question answering, summarization, text classification, relation extraction, and multi-choice questions (Section~\ref{met:instruction dataset}).

Furthermore, we use the off-the-shelf MedCPT~\citep{jin2023medcpt} retriever and construct biomedical corpora as follows: PubMed Abstract, PMC Full Text, Clinical Guideline, and Medical Textbook, all tailored to biomedical and clinical text (Section~\ref{met:biomedical_ret}).
The training process for the self-reflection language model and the domain-specific instruction-tuned language model is similar to Self-RAG, except that, instead of directly training instructions into the LLaMA2~\citep{touvron2023llama2} model, we achieve better performance by training the model weights provided by Self-RAG (Section~\ref{met:critic_lm}, \ref{met:generator_lm}).
The goal of our work is to \textit{construct a language model encoded with domain-specific knowledge, enabling it to autonomously assess explanations and answers it generates}.

\textcolor{redcolor}{
Self-BioRAG demonstrates its effectiveness using five open-domain question-answering (QA) benchmark datasets: multi-choice QA (MedQA~\citep{jin2021disease}, MedMCQA~\citep{pal2022medmcqa}, and MMLU~\citep{hendrycks2020measuring}) and long-form QA (LiveQA~\citep{abacha2017overview} and MedicationQA~\citep{abacha2019bridging}).
Experimental results on the multi-choice QA datasets demonstrate that Self-BioRAG significantly outperforms open-foundation LLMs and RAG approaches with a parameter size of 7B or less.
Self-BioRAG achieves a 7.2\% absolute improvement compared to the state-of-the-art model.
In long-form QA datasets, we notice a substantial difference in the terminologies used despite generating predictions that are similar to answers.
We demonstrate that domain-specific components contribute to the performance gains, with training on domain-specific instructions showing the highest improvement.
Our biomedical corpora supplement scarce knowledge, and particularly, Self-BioRAG uses appropriate documents if needed corresponding to the benchmark datasets.
We further analyze that using reflective tokens to adaptively retrieve factual content is effective in solving open-domain question-answering datasets.
Overall, Self-BioRAG finds the clues in the question, retrieves relevant evidence, and understands how to answer with information using encoded knowledge.
}

Our \textbf{contributions} are as follows:
(1) We introduce a Self-BioRAG framework which is extensively trained on biomedical and clinical instructions.
(2) We prove that domain-specific components such as retriever, documents, and instruction sets are necessary to address its domain-related instructions. 
(3) Self-BioRAG demonstrates its effectiveness in three open-domain biomedical question-answering benchmark datasets by achieving an average absolute improvement of 7.2\% compared to the state-of-the-art open-foundation model with a parameter size of 7B or less.
(4) We release our biomedical instruction sets, code for training our components used in Self-BioRAG, and model weights (7B \& 13B) to be more capable in biomedical and clinical domains.
\vspace{-0.5cm}
\begin{figure*}[t]
  \centering
  \includegraphics[width=\textwidth]{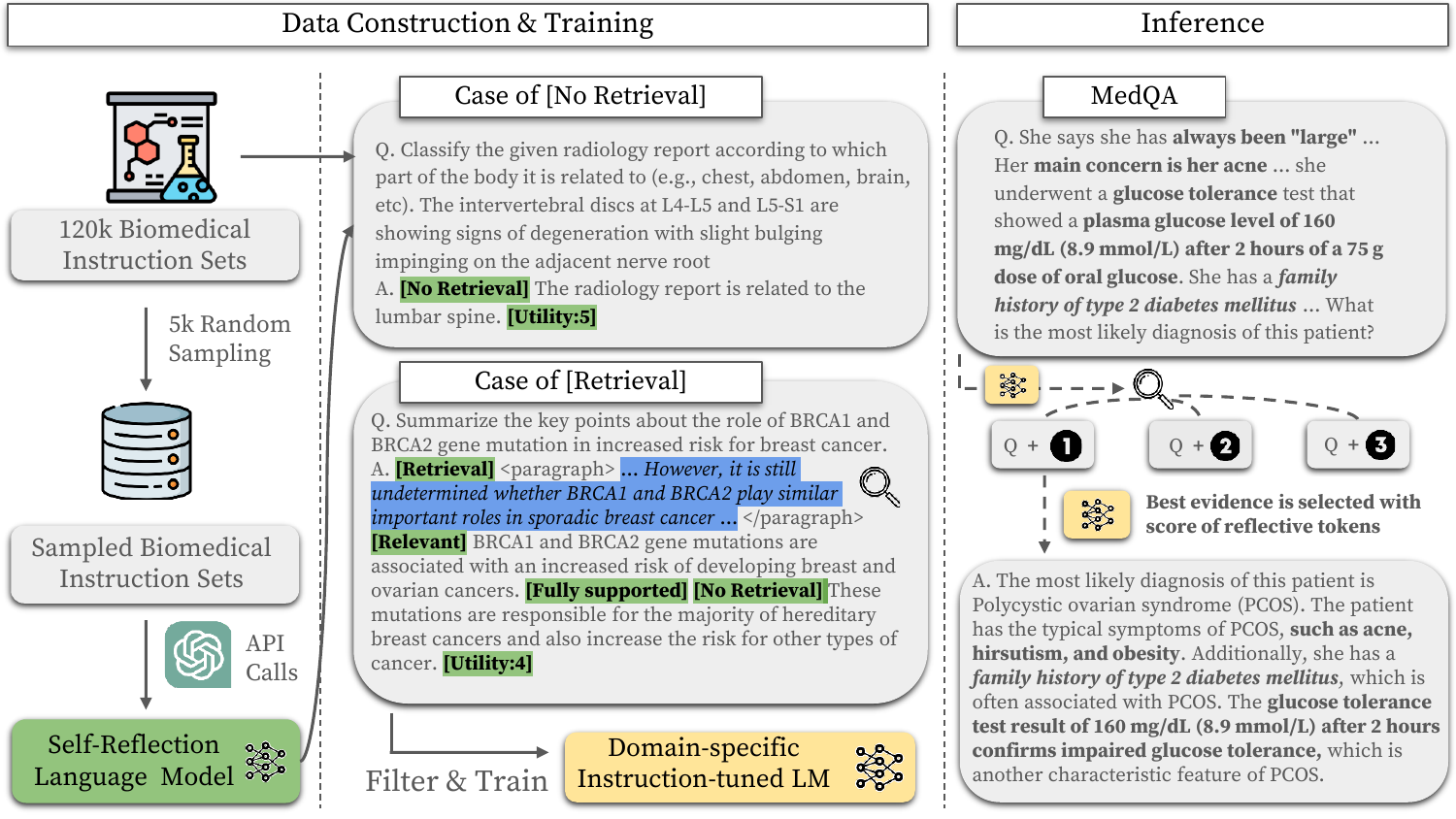}
  \vspace{-0.5cm}
  \caption{
  Overview of our Self-BioRAG process: data construction, training, and inference of Self-Reflection Language Model (critic LM $C$) and Domain-specific Instruction-tuned Language Model (generator LM $M$). We construct 120k biomedical instruction sets using two off-the-shelf instruction sets (Mol-Instructions~\citep{fang2023mol} and MedInstruct~\citep{zhang2023alpacare}) and one self-generated biomedical instruction set. We first sample 5k instructions to generate reflective tokens via GPT-4 API calls and then train the critic LM $C$ with these instructions. Using trained critic LM $C$, we filter out mispredicted reflective tokens, such as [Continue Generation]. We preserve 84k instruction sets annotated with pre-defined reflective tokens to train the generator LM $M$. Note that critic LM $C$ is only used for annotating reflective tokens used to filter instruction sets to train generator LM $M$. After training, the model $M$ can predict whether or not to use the retrieval method and combine the results of evidence and encoded knowledge to answer the question. We use the MedQA~\citep{jin2021disease} test sample to gain a proper understanding of how our Self-BioRAG works.
  }\label{fig:example_figure}
  \vspace{-0.6cm}
\end{figure*}

\section{Background}
\subsection{Proprietary \& Open Language Models}
Instructions serve as guidelines for how language models should perform a particular task.
In the commercial field, proprietary language models such as InstructGPT~\citep{ouyang2022training} and ChatGPT~\citep{openai2023a} have gained significant advantages in tuning through instructions.
However, researchers not involved in commercial fields may face challenges in using these models due to a lack of resources.
Hence, research-friendly open foundation models like the LLaMA family~\citep{touvron2023llama2}, Self-instruct~\citep{wang2022self}, and Alpaca~\citep{taori2023alpaca} are released.
In this regard, domain-specific language models tailored for areas such as biomedical and clinical domains, like Galactica~\citep{taylor2022galactica} and Meditron~\citep{chen2023meditron}, have also been released.
Our research also aims to provide labor-inexpensive methods that are easy to use in various vertical domains, including biomedical and clinical domains.
Specifically, Self-BioRAG strives to develop a model capable of solving challenging tasks, ranging from multi-choice questions to long-form generations.
\vspace{-0.2cm}

\subsection{Learning with Reward Strategy}
The proprietary language models trained with reinforcement learning from human feedback (RLHF), such as ChatGPT~\citep{openai2023a} and GPT-4~\citep{openai2023b}, excel at executing straightforward instructions (e.g., translation, code generation, and question answering) in alignment with human intent~\citep{schulman2017proximal, christiano2017deep, openai2023a, openai2023b, bard2023}.
In Self-RAG~\citep{asai2023self}, a critic language model is employed to offer a cost-effective reward strategy compared to RLHF, utilizing reflective tokens.
The critic model determines whether a given task necessitates retrieval, evaluates the appropriateness of the retrieved context, assesses if the generated rationale aligns with the retrieved context, and ultimately judges the overall utility of the output.
Our Self-BioRAG follows the approach of Self-RAG to create a domain-specific critic language model that not only maintains the aforementioned capabilities but is also well-versed in biomedical text.

\subsection{Retrieval-Augmented Generation}
The retrieval-augmented generation (RAG) significantly enhances performance in knowledge-intensive tasks and open-domain question-answering by providing context as input to the language model~\citep{lewis2020retrieval, mao2021generation, kang2023knowledge}.
The retriever also plays a crucial role in language models by providing evidence for pre-training and few-shot fine-tuning~\citep{guu2020retrieval, izacard2022few}.
With the recent advancements in instruction language models, the combination of retriever and language models involves either using the retriever in advance to fetch evidence or iteratively retrieving it when needed~\citep{jiang2023active, shao2023enhancing}.
Our base framework, Self-RAG~\citep{asai2023self}, deviates from these approaches by being designed to perform retrieval on-demand, resulting in better cost efficiency compared to scenarios where retrieval is always active.
However, in domain-specific fields like biomedical or clinical domains, the general method of retrieving context may not be applicable.
Therefore, Self-BioRAG utilizes retrieval methods and documents tailored to specific domains, retrieving meaningful context that aligns with the intended field.
\vspace{-0.2cm}

\begin{table*}[]
{\resizebox{1.0\textwidth}{!}{
\begin{tabular}{ l c l }
\toprule
\multicolumn{1}{ c }{\textbf{Dataset Name}} & \multicolumn{1}{c}{\begin{tabular}[c]{@{}c@{}}\textbf{\# of instances}\\ \textbf{(Original → Filtered)}\end{tabular}} & \multicolumn{1}{c }{\textbf{Tasks Types}}                                                                 \\ \midrule
Mol-Instructions~\citep{fang2023mol} & 51,493 → 38,156         & Information Extraction, Question Answering, Multi-Choice Question                                  \\ 
MedInstruct~\citep{zhang2023alpacare} & 52,002 → 36,429         & Question Answering, Summarization, Text Classification, Multi-Choice Question                      \\
Biomedical Instructions (Ours) & 18,854 → 10,143         & Text Generation, Question Answering,  Relation Extraction, Text Classification, Summarization  \\ \midrule
Total & 122,349 → 84,728 & Question Answering, Information Extraction, Text Classification, Summarization, Text Generation \\
\bottomrule
\end{tabular}}}{}
\caption{\textcolor{redcolor}{Statistics of biomedical instruction sets. We filter instructions using the critic language model $C$ and use it to train the generator language model $M$.}}
\label{tab:biomedical_instruction_statistics}
\vspace{-0.5cm}
\end{table*}


\section{Self-BioRAG}
In this section, we outline the process of creating our Self-BioRAG framework using various biomedical components.
First, we leverage three datasets consisting of biomedical instructions, which are used to train language models to align with human intentions for biomedical text (Section \ref{met:instruction dataset}).
To supplement scarce knowledge via relevant documents in the biomedical or clinical domains, we employ an off-the-shelf MedCPT~\citep{jin2023medcpt} retriever, known for its effectiveness in retrieving relevant documents in biomedical and clinical domains (Section \ref{met:biomedical_ret}).
Subsequently, we develop a critic language model $C$ to annotate the instruction sets which will contain information for facilitating an autonomous assessment of reflective criteria (Section \ref{met:critic_lm}). 
Lastly, we perform training on our generator model $M$ using the instruction sets created with diverse biomedical components (Section \ref{met:generator_lm}).
We depict our processes of data generation, training, and inference in Figure~\ref{fig:example_figure}.

\subsection{Biomedical Instruction Datasets}
\label{met:instruction dataset}
\paragraph{\textbf{List of Instruction Datasets for Biomedical and Clinical Domains.}}
To train the self-reflection language model (LM), also referred to as the critic LM $C$, we utilize diverse text triplets (instruction, input, output).
Specifically, we collect two off-the-shelf instruction sets (Mol-Instructions~\citep{fang2023mol} and MedInstruct~\citep{zhang2023alpacare}), which include tasks like open-generation, true or false, and multi-choice questions.
In addition to the distributed instruction sets, we synthetically generate an additional 18k biomedical and clinical instructions following the Self-Instruct~\citep{wang2022self}.
In total, we construct 120k biomedical instruction sets addressing diverse tasks: information extraction, question answering, and summarization.
For instance, illustrated in Figure~\ref{fig:example_figure}, the example instruction was set to classify the given radiology report according to which part of the body is related and answer with the lumbar spine.
The statistics of biomedical instruction sets are provided in Table~\ref{tab:biomedical_instruction_statistics}. 
Detailed statistics of our generated instruction sets can be found in Appendix~\ref{app:dataset_stats}.

\subsection{Biomedical Retriever}
\label{met:biomedical_ret}
In the fields of biomedical and clinical domains, researchers and doctors addressing challenging issues typically supplement their knowledge with additional information.
Similarly, for a language model to solve problems, it needs to retrieve relevant documents as needed.
To achieve this, we use the off-the-shelf MedCPT~\citep{jin2023medcpt} retriever\footnote{https://github.com/ncbi/MedCPT}, which is contrastively trained on an unprecedented scale of 255M query-article pairs from PubMed search logs.
To retrieve relevant documents, we compile data from four sources: PubMed Abstract\footnote{https://pubmed.ncbi.nlm.nih.gov/}, PMC Full-text\footnote{https://www.ncbi.nlm.nih.gov/pmc/tools/textmining/}, Clinical Guidelines\footnote{A publicly released subset of 35,733 guideline articles from MEDITRON~\citep{chen2023meditron}, extracted from 8 sources: CCO, CDC, CMA, ICRC, NICE, SPOR, WHO, and WikiDoc}, and English Medical Textbooks\footnote{Medical textbooks widely used by medical students and takers of the United States Medical Licensing Examination (USMLE), https://github.com/jind11/MedQA}.
We encode these data offline to make it computationally effective.
The documents are segmented into chunks of 128 words with 32-word overlaps to form evidence following previous works \citep{wang2019multi, karpukhin2020dense}.
We first retrieve top-$k$ ($k$=10, in our case) evidence from each source data (total 4$k$ evidence) and then use the reranking module to obtain the final top-$k$ evidence relevant to the query.
Table~\ref{tab:retrieve_statistics} presents the overall statistics of biomedical corpus and how many documents are indexed.

\begin{table}[]
\centering
\begin{tabular}{l c c c}
\toprule
\multicolumn{1}{c}{\textbf{Data}} & \textbf{\# Documents} & \textbf{\# Chunks} & \textbf{Embedding Size} \\ \midrule
PubMed& 36,533,377 & 69,743,442 & 400GB \\ 
PMC & 1,060,173  & 46,294,271 & 160GB \\ 
CPG & 35,733  & 606,785   & 3.5GB  \\ 
Textbook & 18 & 133,875 & 0.7GB \\
\bottomrule
\end{tabular}
\vspace{0.1cm}
\caption{Statistics of the indexed biomedical corpus. 
CPG stands for Clinical Practice Guideline. 
}
\label{tab:retrieve_statistics}
\vspace{-0.5cm}
\end{table}

\subsection{Self-Reflection Language Model (Critic Language Model)}
\paragraph{\textbf{Data Construction of Critic LM $C$.}}
\textcolor{redcolor}{
We collect a total of 120k biomedical instruction sets and randomly sample 5k examples ($D_s$) to train the critic LM $C$.
We use GPT-4 API Calls to generate reflective tokens $r$, guiding the critic model $C$ in learning how to predict these tokens.
We follow the usage of four types of reflective tokens $r$ employed in Self-RAG, as described in Table~\ref{tab:reflective_tokens}.}
Detailed statistics and prompts used in generating each type of reflective token are provided in Appendix~\ref{app:critic}.
Exploring other reflective tokens suitable for specific domains is left for future work.

\paragraph{\textbf{Process of Training Critic LM $C$.}}
We observe that a critic LM released by Self-RAG~\citep{asai2023self}\footnote{https://github.com/AkariAsai/self-rag} is unsuitable for predicting reflective tokens in biomedical domains\footnote{See Appendix~\ref{app:critic_lm_performance} for a comparison of critic LMs' performances}. 
Thus, we decide to develop a domain-specific critic LM $C$ using our biomedical instruction sets.
We split the sampled instruction sets into train and dev to train and assess the performance of the critic LM $C$.
We train the model using four types of reflective tokens $r$ annotated with GPT-4 API calls.
\textcolor{redcolor}{
We initialize the critic LM $C$ with a pre-trained language model (here we use LLaMA2~\citep{touvron2023llama2}) and train it on the sampled dataset $D_s$ to maximize the likelihood as below.
\begin{equation}
\underset{C}{max}~\mathbb{E}_{(x, y, r) \sim D_{s}} \text{log}~ p_C(r|x,y)
\end{equation}
}
Note that training the critic LM $C$ signifies it to predict pre-defined reflective tokens given instruction, output, and optionally evidence.
We use trained LM $C$ to annotate whole instruction sets and filter out instances when it mispredicts the reflective tokens that are not pre-defined such as [Continue Generation].
We provide detailed hyperparameters used to train the critic LM $C$ in Appendix~\ref{app:training_details}.

\paragraph{\textbf{Annotating Biomedical Instruction Sets Using Critic LM $C$.}}
After training, the model $C$ predicts four types of reflective tokens: (1) identifying whether a question requires retrieval (RET); (2) determining if retrieved evidence provides useful information to solve a question (REL); (3) assessing whether all statements of answers can be supported by evidence (SUP); (4) evaluating whether all statements of answers are a useful response to the question (USE).
For example, in Figure~\ref{fig:example_figure}, the model $C$ predicts the retrieval of factual content related to the role of BRCA1 and BRCA2 gene mutation ([Retrieval]).
Then, the model predicts that the retrieved evidence provides a fact that BRCA1 and BRCA2 play similar roles in breast cancer and sporadic cancer ([Relevant]).
By comparing a statement of the answer and retrieved evidence, the model $C$ predicts that the answer could be supported by evidence ([Fully supported]).
Finally, the model $C$ suggests that all statements of answers are useful responses to the question ([Utility:4]).
After annotating each type of reflective token, we aggregate all results to construct a complete instance as above.
We provide detailed instructions to annotate the biomedical instruction dataset using the critic LM in Appendix~\ref{app:examples_train_generator}.

\label{met:critic_lm}
\begin{table}[t]
{\resizebox{1.0\columnwidth}{!}{
\begin{tabular}{ l l l l }
\toprule
\multicolumn{1}{c}{\textbf{Type}} & \textbf{Input} & \multicolumn{1}{c}{\textbf{Output}}    & \multicolumn{1}{c }{\textbf{Definitions}} \\ \midrule
RET    & $x / x, y$   & \{yes, no, continue\} & Decides when to retrieve using $R$  \\
REL    & $x, e$  & \{\textbf{relevant}, irrelevant\} & $e$ provides useful information to solve $x$       \\
SUP    & $x, e, y$    & \begin{tabular}[c]{@{}l@{}}\{\textbf{fully supported}, partially\\ supported, no support\}\end{tabular} & \begin{tabular}[c]{@{}l@{}}All of the verification-worthy statement \\ in $y$ is supported by $e$ \end{tabular} \\
USE    & \multicolumn{1}{l}{$x, y$} & \{\textbf{5}, 4, 3, 2, 1\}   & $y$ is a useful response to $x$     \\ \bottomrule
\end{tabular}
}}{}
\vspace{0.1cm}
\caption{\textcolor{redcolor}{Reflective tokens $r$ used in Self-BioRAG. $x$, $y$, and $e$ respectively indicate input, output, and evidence. Specific reflective tokens highlighted in bold are desirable during data construction as they contribute to preserving the existing instruction data possible.}}
\label{tab:reflective_tokens}
\vspace{-0.5cm}
\end{table}

\subsection{Domain-Specific Instruction-Tuned Language Model (Generator Language Model)}
\label{met:generator_lm}
\paragraph{\textbf{Data Construction Using Critic LM $C$ \& Training Generator LM $M$.}}
We use MedCPT to retrieve top-$k$ evidence following an instruction that necessitates retrieval of biomedical context.
After retrieving relevant documents, we use the critic LM $C$ to predict each reflective token as described in Table~\ref{tab:reflective_tokens}.
Consequently, we preserve 84k filtered instances of biomedical instruction sets annotated with pre-defined reflective tokens, instruction, and output triplets to train generator LM $M$.
We want to point out that the critic LM $C$ is only used to annotate reflective tokens to generate biomedical instruction sets to train generator LM 
$M$.
\textcolor{redcolor}{
We fine-tune these filtered 84k biomedical instructions on the generator model to predict answer with reflective tokens as below,
\begin{equation}
    \underset{M}{max}~\mathbb{E}_{(x, y, r) \sim D} ~\text{log}~ p_M(y, r|x)
\end{equation}
where $D$ stands for filtered instruction sets annotated with pre-defined reflective tokens $r$.}
This enhances generalizability in the biomedical and clinical domains preserving the abilities of text generation and self-assessment of its generated explanations with reflective tokens.

\paragraph{\textbf{Inference Process of Self-BioRAG.}}
In Figure~\ref{fig:example_figure}, we present a MedQA~\citep{jin2021disease} example to illustrate our Self-BioRAG inference offline.
For instance, the question is inquiring about the diagnosis of a female patient exhibiting symptoms of obesity, acne, and has a history of type 2 diabetes mellitus. 
The generator model $M$ determines the need to retrieve a relevant document and selects the best evidence from the top-$k$ retrieved documents based on a score $S$, calculated as the weighted sum of reflective tokens, using the same hyperparameters as Self-RAG,
\begin{align*} 
S(Critique) &= \sum_{G \in \mathbb{G}} w^{G}s^{G}, ~~ \mathbb{G} = {REL \cup SUP \cup USE}  \\ 
s^{G} &=  \frac{p(\hat{r})}{\sum_{i=1}^{N^{G}} p(r_i)}
\end{align*}
where $s^G$ denotes the generation probability of the most desirable reflective token $\hat{r}$ (e.g., [Fully supported]) for reflective token type $G$ (e.g., SUP) and $w^G$ represents the hyperparameter providing weight for $s^{G}$. 
We can set the weight $w^G$ to adjust our behavior at inference time.
For example, to find the most relevant document $e$ related to question $x$, we can set a weight term $REL$ score higher.
Self-BioRAG is tailored to conditionally generate text without any additional training which could need balancing the trade-off between multiple preferences~\citep{touvron2023llama2, wu2023fine}. 

The prioritized evidence includes information on \textit{the family history of type 2 diabetes mellitus} and \textit{the patient's diagnosis of polycystic ovarian syndrome (PCOS)}.
Due to space limitations, we display partial information in the figure; please refer to the complete case in Table~\ref{tab:case_report}.
Consequently, the generator model $M$ generates the following text: (1) the patient has acne and obesity, typical symptoms of PCOS; (2) the patient has a family history of type 2 diabetes mellitus, often associated with PCOS; (3) the patient underwent a glucose tolerance test, a characteristic feature of PCOS.
Overall, our Self-BioRAG identifies clues in the question, retrieves factual content if necessary, and responds with encoded knowledge, similar to how doctors approach such cases.

\section{Experimental Details}

\begin{table*}[t]
\centering
{\resizebox{0.8\textwidth}{!}{
\begin{tabular}{l cccc  c}
\toprule
 & & \multicolumn{4}{c}{\textbf{Open-domain Biomedical Benchmark}}     \\ 
 \cmidrule{3-6}
 \multicolumn{1}{c}{\begin{tabular}[c]{@{}c@{}}\textbf{Model}\end{tabular}} & \multicolumn{1}{c}{\begin{tabular}[c]{@{}c@{}}\textbf{Params.}\end{tabular}}& \multicolumn{1}{c}{\begin{tabular}[c]{@{}c@{}}\textbf{MedQA}\\ \textbf{(Acc.)}\end{tabular}} & \multicolumn{1}{c}{\begin{tabular}[c]{@{}c@{}}\textbf{MedMCQA}\\ \textbf{(Acc.)}\end{tabular}} & \multicolumn{1}{c}{\begin{tabular}[c]{@{}c@{}}\textbf{MMLU-Med}\\ \textbf{(Acc.)}\end{tabular}} & \textbf{Average} \\ \midrule
 \textit{Proprietary LM} \\
 \midrule
Med-PaLM \citep{singhal2022large} & \multicolumn{1}{c}{540B} & \multicolumn{1}{c}{60.3}  & \multicolumn{1}{c}{56.5}  & \multicolumn{1}{c}{75.6}   & 64.1   \\
GPT-3.5 \citep{openai2023a} & \multicolumn{1}{c}{-} & \multicolumn{1}{c}{53.6}  & \multicolumn{1}{c}{51.0}   &  \multicolumn{1}{c}{67.3}     &  57.2  \\
GPT-4-base \citep{openai2023b} & \multicolumn{1}{c}{-} & \multicolumn{1}{c}{86.1}  & \multicolumn{1}{c}{73.7}   &  \multicolumn{1}{c}{89.9}     & 83.2   \\ 
\midrule
\textit{Open LM} \\
\midrule
Alpaca~\citep{taori2023alpaca} & 7B    & \multicolumn{1}{c}{23.6} & \multicolumn{1}{c}{30.4}    & \multicolumn{1}{c}{34.4}      &  29.5  \\
FLAN-T5~\citep{chung2022scaling} & 3B     & \multicolumn{1}{c}{36.1} & \multicolumn{1}{c}{20.0}    & \multicolumn{1}{c}{44.2}       &  33.4  \\ 
PMC-LLaMA~\citep{wu2023pmc} & 7B    & \multicolumn{1}{c}{26.7} & \multicolumn{1}{c}{26.5}    & \multicolumn{1}{c}{25.8}      &  26.3  \\
Galactica~\citep{taylor2022galactica} & 6.7B & \multicolumn{1}{c}{29.5} & \multicolumn{1}{c}{32.7}      & \multicolumn{1}{c}{39.4}     & 33.9   \\
MedAlpaca~\citep{han2023medalpaca} & 7B    & \multicolumn{1}{c}{35.6} & \multicolumn{1}{c}{36.6}    & \multicolumn{1}{c}{39.5}      &  37.2  \\
MEDITRON~\citep{chen2023meditron} & 7B     & \multicolumn{1}{c}{36.5} & \multicolumn{1}{c}{37.1}   & \multicolumn{1}{c}{41.1}     & 38.3   \\
LLaMA2~\citep{touvron2023llama2} & 7B     & \multicolumn{1}{c}{35.2} & \multicolumn{1}{c}{36.3}     & \multicolumn{1}{c}{46.3}     & 39.3  \\
   \midrule
\textit{Open LM (LLaMA2) + Retrieval} \\
\midrule
RAG   & 7B & \multicolumn{1}{c}{36.2} & \multicolumn{1}{c}{38.3}   & \multicolumn{1}{c}{47.7}     & 40.7   \\
Self-RAG~\citep{asai2023self}   & 7B & \multicolumn{1}{c}{31.2} & \multicolumn{1}{c}{36.5}   & \multicolumn{1}{c}{45.7}     & 37.8   \\
Self-BioRAG (Ours)   & 7B & \multicolumn{1}{c}{\textbf{43.6}} & \multicolumn{1}{c}{\textbf{42.1}}      & \multicolumn{1}{c}{\textbf{53.9}}     &  \textbf{46.5}   \\ 
Self-BioRAG (Ours)   & 13B & \multicolumn{1}{c}{\textbf{48.6}} & \multicolumn{1}{c}{\textbf{44.0}}      & \multicolumn{1}{c}{\textbf{57.2}}     &  \textbf{49.9}   \\
 \bottomrule
\end{tabular}}}{}
\vspace{0.2cm}
\caption{\textcolor{redcolor}{Experimental results on biomedical benchmark datasets. We use 3-shot examples as guidelines for language models to address benchmark instances.
These examples are chosen from each training dataset using k-nearest-neighbor~\citep{guo2003knn}. 
Since the MMLU dataset lacks training data, we employ the same examples detailed in the appendix of MedPALM~\citep{singhal2022large}.
The score of GPT-3.5 and GPT-4-base models are from the following paper~\citep{nori2023capabilities}.
We use biomedical corpus (e.g., PubMed, PMC, CPG, and Textbook) as evidence during inference on the RAG setting.
The best score is highlighted in bold for the parameter size of 7B or less and our 13B model.}
}
\label{tab:main_table}
\vspace{-0.5cm}
\end{table*}

\paragraph{\textbf{Benchmark Evaluations.}} 
We evaluate the downstream task performance of Self-BioRAG fine-tuned on Self-RAG.
\textcolor{redcolor}{
Our experiments encompass multi-choice and long-form question-answering benchmarks designed to gauge biomedical and clinical knowledge encoded in language models. 
Following the approach of previous work, such as MedPaLM~\citep{singhal2022large}, we use three datasets for few-shot evaluation on open-domain question-answering: MedQA~\citep{jin2021disease}, MedMCQA~\citep{pal2022medmcqa}, and MMLU~\citep{hendrycks2020measuring}.
In MMLU, we extract six clinical topics: anatomy, clinical knowledge, college biology, college medicine, medical genetics, and professional medicine.
For long-form question-answering benchmarks, we use two datasets for evaluation: LiveQA~\citep{abacha2017overview} and MedicationQA~\citep{abacha2019bridging}.
}

\paragraph{\textbf{Baselines.}}
In Table~\ref{tab:main_table}, we compare Self-BioRAG with proprietary, open foundation, and open foundation with retrieval-augmented language models.
We report the score of the proprietary language model as presented in Med-PaLM~\citep{singhal2022large} and the GPT-3.5 and GPT-4-base scores as presented in \cite{nori2023capabilities} to establish the upper bound of benchmark datasets (Row 1-3).
\textcolor{redcolor}{Open foundation models, pre-trained for sequence-to-sequence generation with instruction tuning, such as Alpaca~\citep{taori2023alpaca} and Flan-T5~\citep{chung2022scaling}, are reported (Row 4 \& 5), as well as models fine-tuned on the specific vertical domains (e.g., biomedical and clinical), like PMC-LLaMA~\citep{wu2023pmc}, Galactica~\citep{taylor2022galactica}, MedAlpaca~\citep{han2023medalpaca}, and Meditron~\citep{chen2023meditron} (Row 6-9).
LLaMA2~\citep{touvron2023llama2} demonstrates state-of-the-art performance in open-foundation 7B models in our experiment (Row 10). 
Therefore, we employ LLaMA2 for the result of retrieval-augmented generation (RAG) and provide the top-10 evidence collected from the biomedical corpus using the MedCPT retriever (Row 11). 
Due to the length limit of RAG for input, we can only leverage the top 1 evidence in input and few-shot examples.
Additionally, we report Self-RAG~\citep{asai2023self} using Contriever~\citep{izacard2022unsupervised} fine-tuned on MSMARCO~\citep{bajaj2016ms} with the Wikipedia corpus (Row 12).
We compare these baselines with our Self-BioRAG framework which is trained with biomedical components.}

\paragraph{\textbf{Training and Inference Settings.}}
Self-BioRAG is trained with 84k biomedical instruction sets filtered using a trained critic language model (LM).
We adopt the Self-RAG critic LM as our base model and fine-tune it with 5k sampled instruction sets annotated by GPT-4 API calls.
As training on the Self-RAG generator LM yields better results, we fine-tune our biomedical instruction sets instead of training directly on LLaMA2~\citep{touvron2023llama2} or Meditron~\citep{chen2023meditron}.
For the retriever, we use the off-the-shelf MedCPT~\citep{jin2023medcpt} retriever, specialized in retrieving documents based on biomedical queries and retrieving up to ten evidence for each input.

\begin{table}[t]
\centering
{\resizebox{0.4\textwidth}{!}{
\begin{tabular}{ l cc }
\toprule
\multicolumn{1}{c}{\multirow{1}{*}{\textbf{Model}}} &  \begin{tabular}[c]{@{}c@{}}\textbf{LiveQA} \\ \textbf{(R1/R2/RL/BS)}\end{tabular}      & \begin{tabular}[c]{@{}c@{}}\textbf{MedicationQA} \\ \textbf{(R1/R2/RL/BS)}\end{tabular}   \\ \midrule
MEDITRON~\citep{chen2023meditron}    & 5.5 / 0.0 / 2.5 / 77.2     & 4.1 / 0.2 / 3.3 / 75.9  \\ LLaMA2~\citep{touvron2023llama2} & 8.8 / 1.9 / 6.2 / \textbf{78.8} & 5.7 / 1.2 / 4.4 / 77.6  \\
RAG & 11.5 / 2.3 / 11.1 / 69.5 & 9.8 / 1.3 / 4.8 / 72.9 \\
Self-BioRAG (Ours) & \textbf{19.7} / \textbf{3.1} / \textbf{13.4} / 77.2 & \textbf{17.6} / \textbf{3.3} / \textbf{13.5} / \textbf{80.2} \\ \bottomrule
\end{tabular}}}{}
\vspace{0.1cm}
\caption{\textcolor{redcolor}{Results of Long-form question-answering benchmark. 
We report the Rouge-1 (R1), Rouge-2 (R2), Rouge-L (RL) scores to measure n-gram recall performance and report BERTScore (BS) which computes the similarity of two sentences as a sum of cosine similarities between their tokens' embeddings.
}}
\vspace{-0.2cm}
\label{tab:lfqa_result}
\end{table}

For inference, we use vllm~\citep{kwon2023efficient} to speed up our inference time.
Following Self-RAG~\citep{asai2023self}, we assign the same weight terms for reflective tokens (e.g., REL, SUP, USE) in decoding.
We adopt adaptive retrieval by default which dynamically decides when to retrieve the evidence by predicting a reflective token [Retrieval].
we retrieve the top ten evidence from the biomedical corpus processed offline.
We provide details of the retrieved percentage of source data used to evaluate biomedical benchmark datasets in Section~\ref{ana:source_data}.

\section{Results and Analysis}
\subsection{Experimental Results}
\paragraph{\textbf{What Contributes to the Performance Improvements in Self-BioRAG?}}
\textcolor{redcolor}{
In Table~\ref{tab:main_table}, we compare our Self-BioRAG with open foundation language model (LM) and retrieval augmented generation (RAG).
With a parameter size of 7B or less, our Self-BioRAG outperforms other open foundation LMs (Row 4-10) in all three biomedical benchmark datasets (MedQA, MedMCQA, and MMLU-Med).
We also compare our model with baselines using retrieval evidence.
The RAG pipeline faces two challenges: it struggles to identify crucial evidence and encounters limitations in incorporating numerous pieces of evidence due to constraints on input length.
However, our Self-BioRAG outperforms the RAG baseline and can prioritize important evidence via the values of reflective tokens, which is useful for analyzing all the retrieved evidence (Row 11 \& 13).
Although Self-RAG is fine-tuned on LLaMA2, we observe that Self-RAG cannot generalize to biomedical benchmark datasets, resulting in a performance drop (Row 10 \& 12).
By providing a biomedical critic LM and corpus to train a biomedical generator LM, our Self-BioRAG achieves state-of-the-art performance on 7B parameters in MedQA, MedMCQA, and MMLU-Med datasets.
We also provide the 13B performance of our Self-BioRAG model to demonstrate the effectiveness of our framework works in other model parameters (Row 14).}
We provide the detailed performance of specific MMLU datasets in Appendix~\ref{app:mmlu-med}.

\textcolor{redcolor}{
In Table~\ref{tab:lfqa_result}, we compare our Self-BioRAG with two open foundation LM by measuring n-gram recall performance (Rouge Score~\citep{lin2004rouge}) and similarity of token embeddings between prediction and answer (BERTScore~\citep{zhang2019bertscore}).
We observe that although all foundation models do not generate predictions with the exact same words as the answers (lower Rouge Score), they manage to explain well with words that are as similar as possible (high BERTScore).
However, these scores cannot measure whether a model has generated answers with accurate rationale, how much hallucination occurs, how much it includes crucial claims, or whether it has generated answers fluently.
We leave an investigation about detailed capacities related to long-text generation for future works.
}
We aim to analyze the step-by-step process through which our Self-BioRAG achieves its state-of-the-art performance in the following subsection.


\begin{table}[]
\centering
{\resizebox{0.45\textwidth}{!}{
\begin{tabular}{ l c c c c }
\toprule
\multicolumn{1}{ c }{\textbf{Experiment Detail}} & \begin{tabular}[c]{@{}c@{}}\textbf{MedQA} \\ \textbf{(Acc.)}\end{tabular} & \multicolumn{1}{c }{\begin{tabular}[c]{@{}c@{}}\textbf{MedMCQA} \\ \textbf{(Acc.)}\end{tabular}} & \begin{tabular}[c]{@{}c@{}}\textbf{MMLU-Med} \\ \textbf{(Acc.)}\end{tabular} &
\textbf{Average} \\ \midrule
Self-BioRAG & \textbf{43.6} & \textbf{42.1} & \textbf{53.9} & \textbf{46.5}  \\
\quad$-$ Reflective Tokens  & 42.5 & 41.9 & 51.1 & 45.2 (-1.3)   \\ 
\quad\quad$-$ Biomedical Corpora & 40.7 & 40.7 & 49.3 & 43.6 (-2.9)  \\ 
\quad\quad\quad$-$ MedCPT Retriever & 39.8 & 38.9 & 47.6 & 42.1 (-4.4)  \\ 
\quad\quad\quad\quad$-$ Biomedical Instruction Sets  & 34.8 & 36.4 & 46.4 & 39.2 (-7.3) \\
\bottomrule
\end{tabular}}}{}
\vspace{0.1cm}
\caption{
\textcolor{redcolor}{Effect of each domain-adaptaion component.}
}
\vspace{-0.5cm}
\label{tab:ablation_study}
\end{table}

\begin{figure}[t]
\centering
\includegraphics[width=0.48\textwidth]{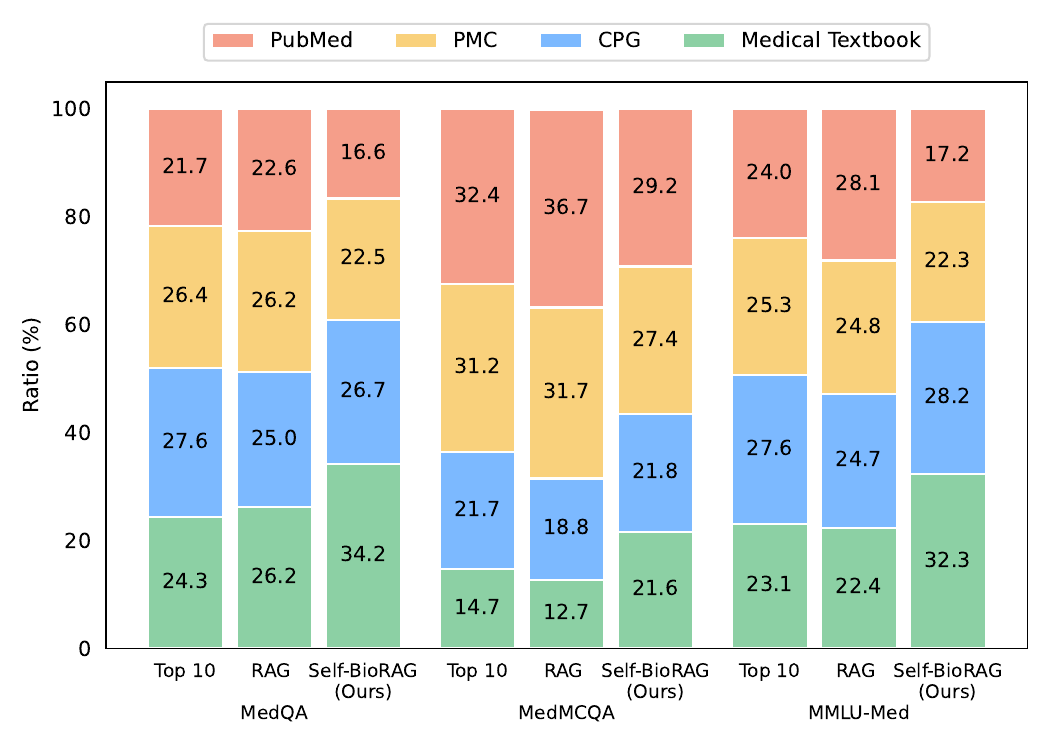}
\vspace{-0.8cm}
\caption{Ratio of retrieved evidences from each of the four biomedical corpora (PubMed, PMC, CPG, Medical Textbook). 
The RAG statistics refer to the top-1 evidence usage ratio, while Self-BioRAG selects the most useful evidence from the top-10 retrieved evidence.}
\label{fig:retrieve_ratio}
\end{figure}

\subsection{Analysis}
\label{ana}
\paragraph{\textbf{Which Domain-Adaptation Components Show the Improvements Compared to Self-RAG?}}
\textcolor{redcolor}{
In Table~\ref{tab:ablation_study}, each experiment involves sequentially reducing components in Self-BioRAG. 
The goal is to identify the factors that significantly contributed to the performance improvement, ultimately leading to the final performance of Self-BioRAG.
First, the controllable generation using reflective tokens affects the rationale which leads to predicting an answer (Row 2).
Then, we observe that using four biomedical corpora (PubMed, PMC, CPG, and Medical Textbook) to retrieve appropriate evidence shows performance improvement compared to Wikipedia evidence (Row 3).
We also use domain-specific MedCPT retriever instead of the Contriever~\citep{izacard2022unsupervised} fine-tuned on MSMARCO~\citep{bajaj2016ms} (Row 4).
Ultimately, the most effective approach was the collection and processing of biomedical instruction sets to create both a critic language model and a generation language model (Row 5). 
We recommend readers collect their domain-specific instructions to address corresponding instructions.
}

\paragraph{\textbf{In Biomedical Corpora, What Evidence is Used to Solve Open-Domain Question-Answering Benchmarks?}}
\label{ana:source_data}
In Figure~\ref{fig:retrieve_ratio}, we compare the ratio of retrieved evidence using the MedCPT~\citep{jin2023medcpt} retriever on four biomedical corpora (PubMed, PMC, CPG, and Medical Textbook).
Even though the index sizes of Medical Textbook and CPG are much smaller than PubMed or PMC, retrieved evidences show even distribution.
Specifically, our Self-BioRAG only retrieves small portions to solve three datasets (MedQA (12\%), MedMCQA (8\%), and MMLU-Med (11\%)) meaning that these open-domain benchmarks don't require that much evidence than expected.
We depict these portions up to 100\% in Figure~\ref{fig:retrieve_ratio}.
We observe a trend in which Self-BioRAG retrieves a higher proportion of information from the Medical Textbook, similar to the approach used in solving USMLE-style questions.
This is also aligned with previous facts that retrieving documents from Medical Textbook can achieve higher performance in clinical questions~\citep{wang2023augmenting, li2023meddm}.

\begin{table}[t]
\centering
{\resizebox{0.48\textwidth}{!}{
\begin{tabular}{ l c c c c}
\toprule
\multicolumn{1}{ c }{\textbf{Methods}} & \multicolumn{1}{c }{\begin{tabular}[c]{@{}c@{}}\textbf{MedQA} \\ \textbf{(Acc.)}\end{tabular}} & \begin{tabular}[c]{@{}c@{}}\textbf{MedMCQA} \\ \textbf{(Acc.)}\end{tabular}  & \begin{tabular}[c]{@{}c@{}}\textbf{MMLU-Med} \\ \textbf{(Acc.)}\end{tabular} & \textbf{Average}\\ \midrule
Only [No Retrieval]    & 39.7 & 41.9 & 52.8 & 44.8   \\
Only [Retrieval]  & 40.1 & \textbf{47.2} & 51.3  & 46.2  \\ 
Adaptive Retrieval (Ours) & \textbf{43.6} & 42.1 & \textbf{53.9} & \textbf{46.5}   \\ 
\bottomrule
\end{tabular}}}{}
\vspace{0.1cm}
\caption{
Effect of adaptive retrieval in Self-BioRAG.
`Only [No Retrieval]' refers to not retrieving any evidence, while `Only [Retrieval]' refers to forcing the retrieval of top 10 evidences.  
}
\vspace{-0.2cm}
\label{tab:use_retrieval}
\end{table}

\paragraph{\textbf{Does Evidence Truly Help to Supplement Limited Knowledge in Self-BioRAG?}}
In Table~\ref{tab:use_retrieval}, we conduct three different experimental settings using open-domain question-answering benchmark datasets:
(1) \textit{No Retrieve}: Answer without utilizing any provided evidence.
(2) \textit{Only Retrieve}: Retrieve top-10 evidence for each question and predict an answer.
(3) \textit{Adaptive Retrieve}: Use a criterion to decide whether to retrieve or not based on the question.
The criterion is set as follows,
\begin{equation}
\frac{p(retrieve = Yes)}{p(retrieve = Yes) + p(retrieve = No)} > \delta.
\end{equation}
where we set $\delta$ hyperparameter as 0.2 for the \textit{Adaptive Retrieve} experiment setting.
Our findings indicate that retrieving relevant documents indeed aids in solving benchmark datasets.
Additionally, we observed that adaptively retrieving shows comparable performance on average with the \textit{Only Retrieve} setting.
This is attributed to the small portion of retrieved evidence used to answer the questions.
While the \textit{Only Retrieve} setting exhibits a substantial improvement in MedMCQA, it shows a performance drop in MMLU-Medical datasets compared to the \textit{No Retrieve} setting, indicating its instability.
As a result, we recommend readers to use the adaptive retrieval setting.

\begin{figure}[t]
\centering
\includegraphics[width=0.45\textwidth]{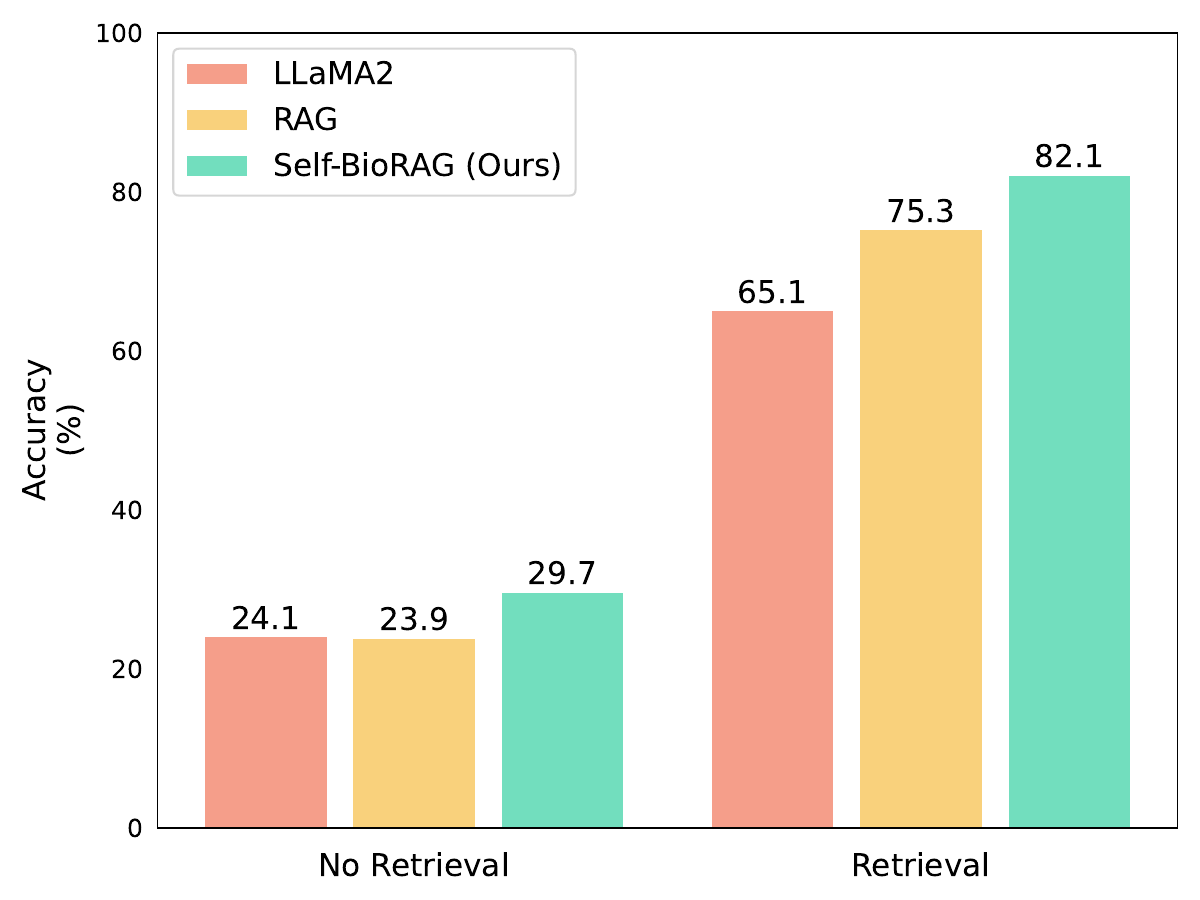}
\vspace{-0.5cm}
\caption{
Performance of LLaMA2, RAG, and Self-BioRAG on examples split into [No Retrieval] and [Retrieval] based on Self-BioRAG using the MedQA test dataset.
}
\vspace{-0.5cm}
\label{fig:selfbiorag_retrieve_noretrieve}
\end{figure}

\begin{table*}[]
{\resizebox{0.95\textwidth}{!}{
\begin{tabular}{ l }
\toprule
\multicolumn{1}{ c }{\textbf{MedQA Dataset}} \\ \midrule
\begin{tabular}[c]{@{}l@{}}Query: A 27-year-old woman presents to the office with \textcolor{bluecolor}{\textbf{concerns about her long struggle with her physical appearance since adolescence}}. She \\ says 
\textcolor{bluecolor}{\textbf{she has always been $``$large$"$}} and was constantly targeted by her classmates and coworkers for being so. \textcolor{exredcolor}{\textbf{Her main concern at the moment is}} \\ \textcolor{exredcolor}{\textbf{her acne and unwanted facial hair}} on her upper lip, for which she often visits a local spa. She has tried numerous diet plans, exercise regimens, and \\ cosmetic 
products with little to no effect. Recently, \textcolor{exredcolor}{\textbf{she underwent a glucose tolerance test that showed a plasma glucose level of 160 mg/dL (8.9 }}\\\textcolor{exredcolor}{\textbf{mmol/L) 
after 2 hours of a 75 g dose of oral glucose}}. \textcolor{bluecolor}{\textbf{She has a family history of type 2 diabetes mellitus}} and a menstrual cycle that occurs every \\45 days. Her pulse is 72/min and the blood pressure is 138/80 mm Hg. On physical examination, her height is \textcolor{bluecolor}{\textbf{160 cm (5 ft 3 in) and her weight is 85}}\\ \textcolor{bluecolor}{\textbf{kg (187 lb)}}, 
and \textcolor{exredcolor}{\textbf{she has severe inflammatory acne over the cheeks and forehead and dark coarse hairs on the back}}. What is the most likely \\diagnosis of this patient? \\
Option A: Hypothyroidism \\
Option B: Idiopathic hirsutism \\ 
Option C: Polycystic ovarian syndrome (PCOS) \\ 
Option D: Ovarian hyperthecosis \\ \\ 
Evidence: \textit{Early Clinical Expressions of Insulin Resistance: The Real Enemy to Look For. Today, a very common clinical scenario is a 17-year}\\\textit{-old \textcolor{bluecolor}{\textbf{female with a family history of type 2 diabetes mellitus (T2DM) and hypertension in her mother}} and two first-degree relatives. Three years ago} \\ \textit{she was \textcolor{bluecolor}{\textbf{diagnosed with polycystic ovarian syndrome (PCOS)}}. She is \textcolor{bluecolor}{\textbf{overweight with her body weight increasing 10 kg}} in the last 2 years. Addition-}\\
\textit{ally, she has acanthosis nigricans (AN) in the axillae and knuckles. Blood tests were unremarkable with a fasting plasma glucose value of 82 mg/dl} \\
\textit{and an HbA1c of 5.0\%.} \\ \\
Prediction: The most likely diagnosis of this patient is Option C: Polycystic ovarian syndrome (PCOS).  \textcolor{bluecolor}{\textbf{The patient has the typical symptoms of}}\\ \textcolor{bluecolor}{\textbf{PCOS}}, \textcolor{exredcolor}{\textbf{such as acne, hirsutism}}, and \textcolor{bluecolor}{\textbf{obesity}}. Additionally, \textcolor{bluecolor}{\textbf{she has a family history of type 2 diabetes mellitus}}, which is often associated with\\ PCOS. \textcolor{exredcolor}{\textbf{The glucose tolerance test result of 160 mg/dL (8.9 mmol/L) after 2 hours confirms impaired glucose tolerance}}, which is another charac-\\teristic feature of PCOS.
\end{tabular} \\ \bottomrule
\end{tabular}}}{}
\vspace{0.1cm}
\caption{Case report of Self-BioRAG prediction using evidence in MedQA dataset. Retrieved evidence is written in \textit{italics}. 
\textcolor{bluecolor}{\textbf{Blue-colored text}} comprises segments connected to key information from retrieved evidence, while \textcolor{exredcolor}{\textbf{red-colored text}} consists of segments tied to the model's parametric knowledge.
}
\label{tab:case_report}
\vspace{-0.5cm}
\end{table*}

\paragraph{\textbf{Distinguishing When to Retrieve Documents in Self-BioRAG.}}
In Figure~\ref{fig:selfbiorag_retrieve_noretrieve}, we evaluate the performance of LLaMA2, RAG (LLaMA2 with MedCPT and biomedical corpora), and Self-BioRAG on examples predicted as [No Retrieval] and [Retrieval] by Self-BioRAG.
To show an overall trend, we use the MedQA dataset here and the rest of the two datasets are in Appendix~\ref{app:medmcqa_mmlu-med}.
Notably, Self-BioRAG retrieves small portions to solve three biomedical benchmarks.
Still, the results demonstrate that Self-BioRAG consistently outperforms other baselines, whether or not retrieved evidence is used.
In situations where retrieval is not necessary (left column), Self-BioRAG $>$ RAG $\approx$ LLaMA2.
The overall trend in the retrieved situation (right column) indicates Self-BioRAG $>$ RAG $\geq$ LLaMA2.
Intuitively, we identify that Self-BioRAG distinguishes well on situations to use retrieved evidence or not depending on questions.


\paragraph{\textbf{Case Report of Using Retrieved Evidence.}}
\label{sec:ana_case_report}
In Table~\ref{tab:case_report}, we present an example from the MedQA dataset to illustrate how Self-BioRAG works.
For instance, a patient exhibits symptoms of physical appearance changes, acne, and a family history of type 2 diabetes mellitus (T2DM).
Self-BioRAG determines the need to retrieve relevant documents containing information on a female diagnosed with polycystic ovarian syndrome (PCOS) and similar symptoms (e.g., T2DM and obesity).
Self-BioRAG determines the patient's diagnosis as PCOS by integrating all three: patient's symptoms, retrieved evidence, and parametric knowledge.
Throughout the query, evidence, and prediction, we color-code using \textcolor{bluecolor}{blue} and \textcolor{exredcolor}{red} to distinguish two categories of related snippets: 1) key information extracted from retrieved evidence and 2) the model's essential parametric knowledge, both of which are leveraged by Self-BioRAG to address the given problem.
Overall, Self-BioRAG finds the clues in the question, retrieves relevant documents if needed, and understands how to answer with information of retrieved evidence and encoded knowledge same as a medical expert would do.
\vspace{-0.6cm}

\section{Conclusion}
In this manuscript, we introduce the Self-BioRAG framework, enabling a Self-RAG~\citep{asai2023self} to generalize to biomedical and clinical domains of instructions.
This framework enhances the generation capacity, facilitates the retrieval of factual content on demand, and enables self-assessment of generated rationales. 
\textcolor{redcolor}{
Our experimental results cover five open-domain question-answering (QA) datasets widely used in biomedical and clinical domains.
In multi-choice QA datasets, Self-BioRAG achieves a 7.2\% absolute improvement compared to the state-of-the-art model among the open foundation 7B models.
In Long-form QA datasets, Self-BioRAG exhibits notable variations in term usage, despite producing predictions that closely resemble answers.}
We demonstrate the necessity of domain-specific components, such as retriever, domain-related document corpus, self-reflection model, and generator model, to address domain-related instructions.
We provide diverse analyses: (1) Self-BioRAG retrieves a larger portion of evidence from Medical Textbook than other corpora to solve USMLE-style questions; (2) Self-BioRAG can distinguish when to retrieve evidence depending on instruction and question; (3) Provided evidence from biomedical corpora genuinely helps supplement scarce knowledge.
\textcolor{redcolor}{
In future works, we aim to explore generating long-form text in a fine-grained evaluation which could interpret how open foundation models (with or without domain adaptation) generate.
}
\vspace{-0.4cm}

\section*{Acknowledgements}
We thank Gangwoo Kim, Hyeon Hwang, Chanhwi Kim, and Akari Asai for the valuable feedback on our work.
This work was supported in part by the National Research Foundation of Korea [NRF-2023R1A2C3004176, NRF-2022R1C1C1008074], the Ministry of Health \& Welfare, Republic of Korea [HR20C0021(3), HR22C1302], the Ministry of Science and ICT (MSIT) [RS-2023-00262002, RS-2022-00155911 (Artificial Intelligence Convergence Innovation Human Resources Development (Kyung Hee University))], and the ICT Creative Consilience program through the Institute of Information \& Communications Technology Planning \& Evaluation (IITP) grant funded by the MSIT [IITP-2024-2020-0-01819].

\bibliographystyle{natbib}
\bibliography{ref}


\appendix
\section{Statistics of our generated Biomedical Instruction Set}
\label{app:dataset_stats}
\begin{figure}[]
\centering              \includegraphics[width=0.235\textwidth]{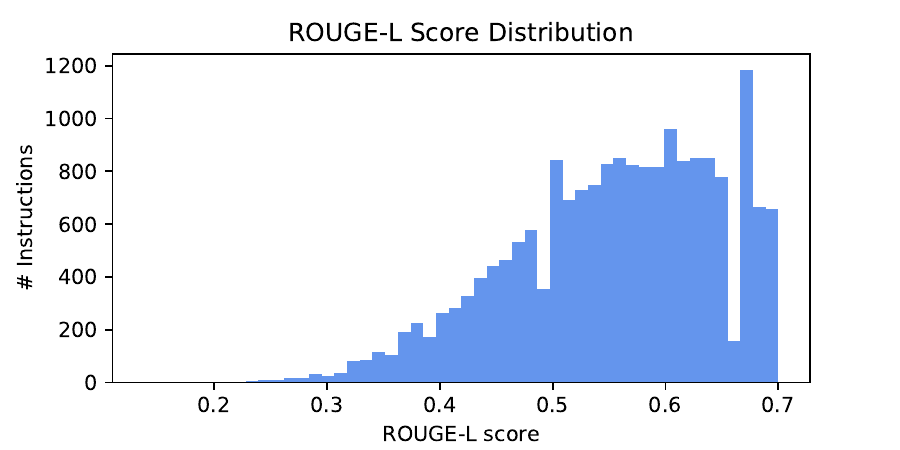}
\includegraphics[width=0.235\textwidth]{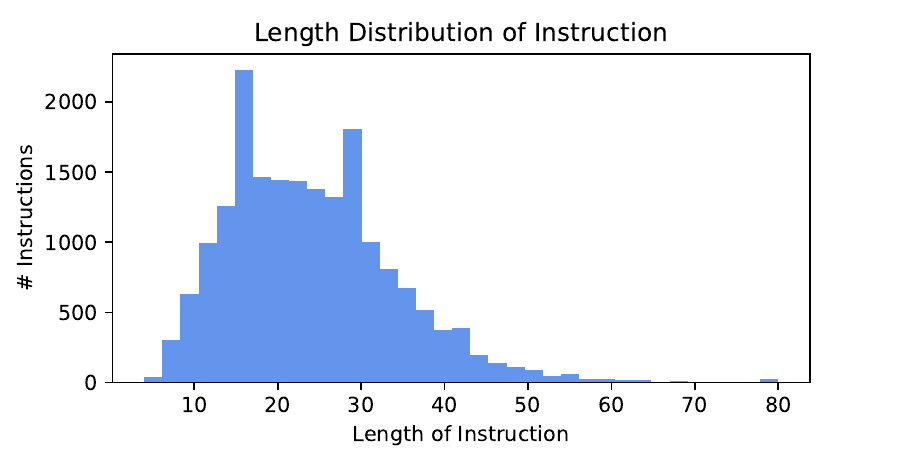}
\includegraphics[width=0.235\textwidth]{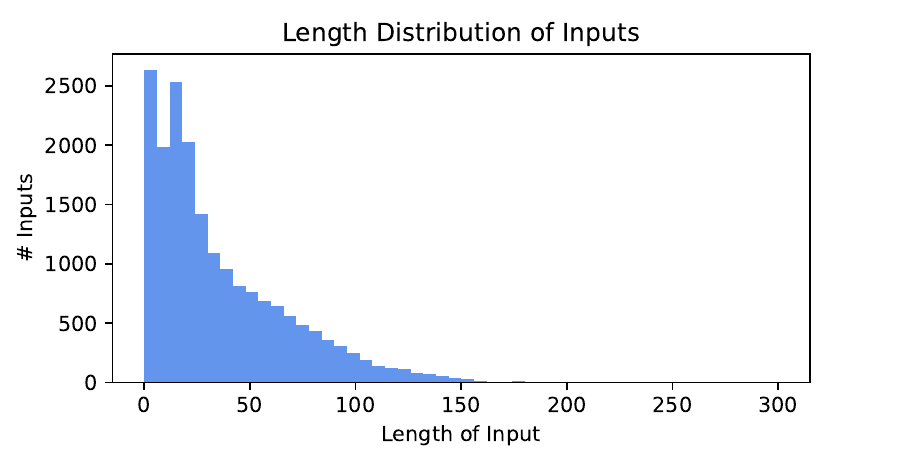}
\includegraphics[width=0.235\textwidth]{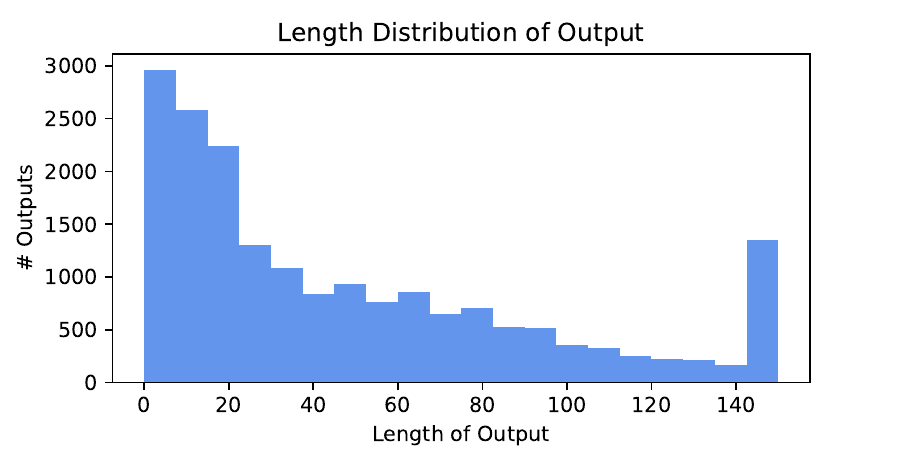}
  \caption{(a) Distribution of ROUGE-L scores between generated instructions and their top-1 similar biomedical seed task instruction. (b) Length distribution of generated instructions. (c) Length distribution of generated inputs. (d) Length distribution of generated outputs. We set the max length of the generated output length as 150.
  }
  \label{fig:distribution_of_dataset}
\end{figure}

In Figure~\ref{fig:distribution_of_dataset}, we depict a ROUGE-L score distribution and show that more various instructions were generated relative to the biomedical seed tasks. 
We compute the ROUGE-L overlap score with the seed tasks' instructions to prevent generating similar instructions as much as possible.
Additionally, we observe that the generated instruction, input, and output triplets show diverse length distribution depending on the condition of instruction.
In Table~\ref{tab:bit_statistics}, we provide other statistics such as the average length of instructions, inputs, and outputs.

\begin{figure}[]
\centering
    \includegraphics[width=0.48\textwidth]
    {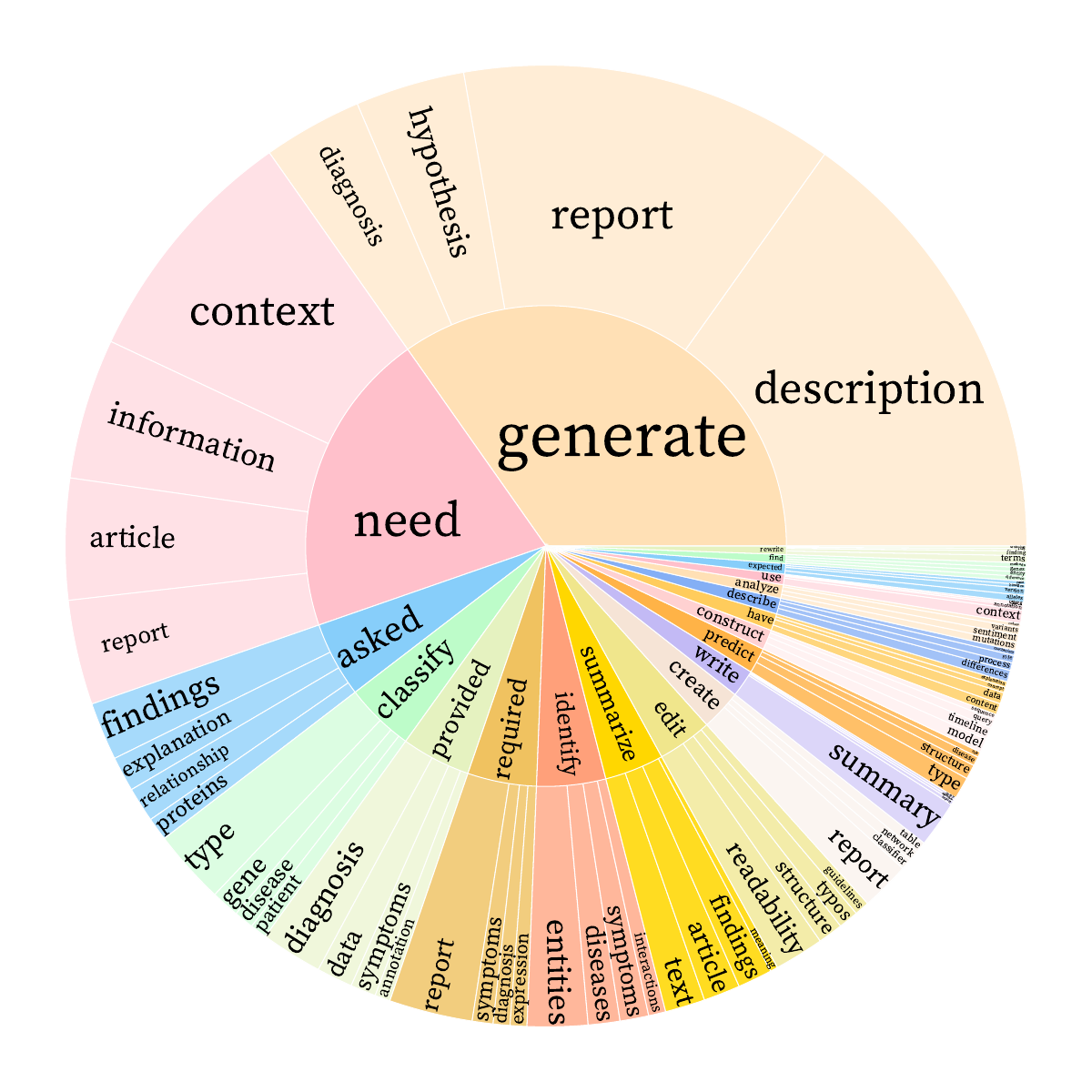}
    \caption{Nested piechart of our biomedical instructions root verbs (inner circle) and their four noun objects (outer circle). It signifies the diversity of generated instructions and relates to biomedical terms such as hypothesis, proteins, diagnosis, and symptoms.}
    \label{fig:pie_chart}
\end{figure}

To guarantee the diverse range of instruction types, we analyze the biomedical instructions using Berkeley Neural Parser\footnote{https://parser.kitaev.io/}~\citep{kitaev2018constituency, kitaev-etal-2019-multilingual}.
We parse the generated instructions and extract the top 20 root verbs in instructions and their direct noun objects.
Due to the requirement of the instruction-generation process, our instruction is entirely related to biomedical terms such as report, hypothesis, and diagnosis.
We depict this information in Figure~\ref{fig:pie_chart}.
Additionally, we provide a prompt to contain a requirement of generating biomedical/medical/clinical topics of instruction.
We random sample 100 instances to manually measure the category of instruction topics.
Our instruction dataset covers various biomedical topics such as text generation (42\%), classification (15\%), symptom identification (7\%), risk value prediction (6\%), clinical note editing (5\%), etc.

\begin{table}[h]
\centering
\begin{tabular}{ l c }
\toprule
\textbf{Statistics} & \\ \midrule
\# of total instances & 18,854 \\
avg. instruction length (words) & 24.2 \\ 
avg. input length (words)       & 35.4 \\
avg. output length (words)      & 47.8 \\
\bottomrule
\end{tabular}
\vspace{0.2cm}
\caption{Statistics of our generated Biomedical Instruction Set.}
\label{tab:bit_statistics}
\vspace{-0.5cm}
\end{table}

\section{Comparison of Critic LM's Performance}
\label{app:critic_lm_performance}
We split 5k instruction sets into train and dev for training and evaluating critic language model performance.
In Table~\ref{tab:critic_lm_performance}, we use GPT-4~\citep{openai2023b} as ground-truth predictions and identify how the critic language model of Self-RAG~\citep{asai2023self} and Self-BioRAG performs.
Due to the uncommon domain of questions, Self-RAG exhibits lower performance compared to our Self-BioRAG.
The critic model of Self-BioRAG trained on Self-RAG predicts reflective tokens of utility and relevance well.
We find that predicting reflective tokens of retrieval and supportiveness shows far lower than we expected.
However, we want to note that distilling knowledge of proprietary language models such as GPT-4 to LLaMA2~\citep{touvron2023llama2} can perform well in predicting reflective tokens.

\begin{table}[h]
\centering
{\resizebox{0.98\columnwidth}{!}{
\begin{tabular}{ l c c c c c }
\toprule
\multicolumn{1}{ c }{\textbf{Model}} & \textbf{Params.} & \begin{tabular}[c]{@{}c@{}}\textbf{Utility}\\ \textbf{(Acc.)}\end{tabular} & \begin{tabular}[c]{@{}c@{}}\textbf{Retrieval}\\ \textbf{(Acc.)}\end{tabular} & \begin{tabular}[c]{@{}c@{}}\textbf{Relevant}\\ \textbf{(Acc.)}\end{tabular} & \begin{tabular}[c]{@{}c@{}}\textbf{Supportive}\\ \textbf{(Acc.)}\end{tabular} \\ \midrule
Self-RAG~\citep{asai2023self}       & 7B     & 78.2 & 65.8 & 88.5 & 69.9 \\ 
Self-BioRAG (Ours)    & 7B     & 89.3 & 73.1 & 91.2 & 75.9 \\ 
Self-BioRAG (Ours)   & 13B    & 91.2 & 75.3 & 91.8 & 77.2  \\ \bottomrule
\end{tabular}}}{}
\vspace{0.2cm}
\caption{Critic LM performance on dev set using GPT-4~\citep{openai2023b} predictions as ground-truth predictions.}
\label{tab:critic_lm_performance}
\end{table}

\begin{figure*}[]
\centering
    \includegraphics[width=0.485\textwidth]{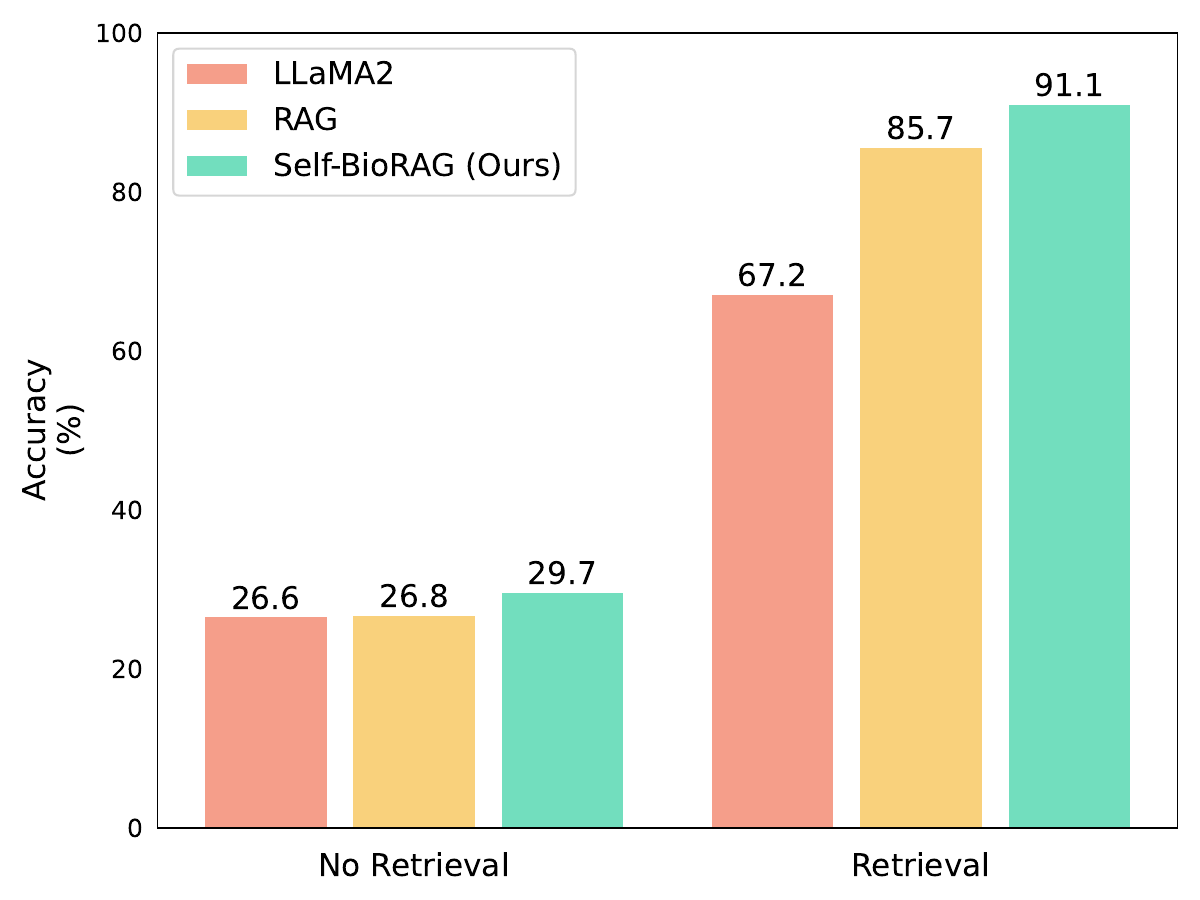}
    \includegraphics[width=0.485\textwidth]{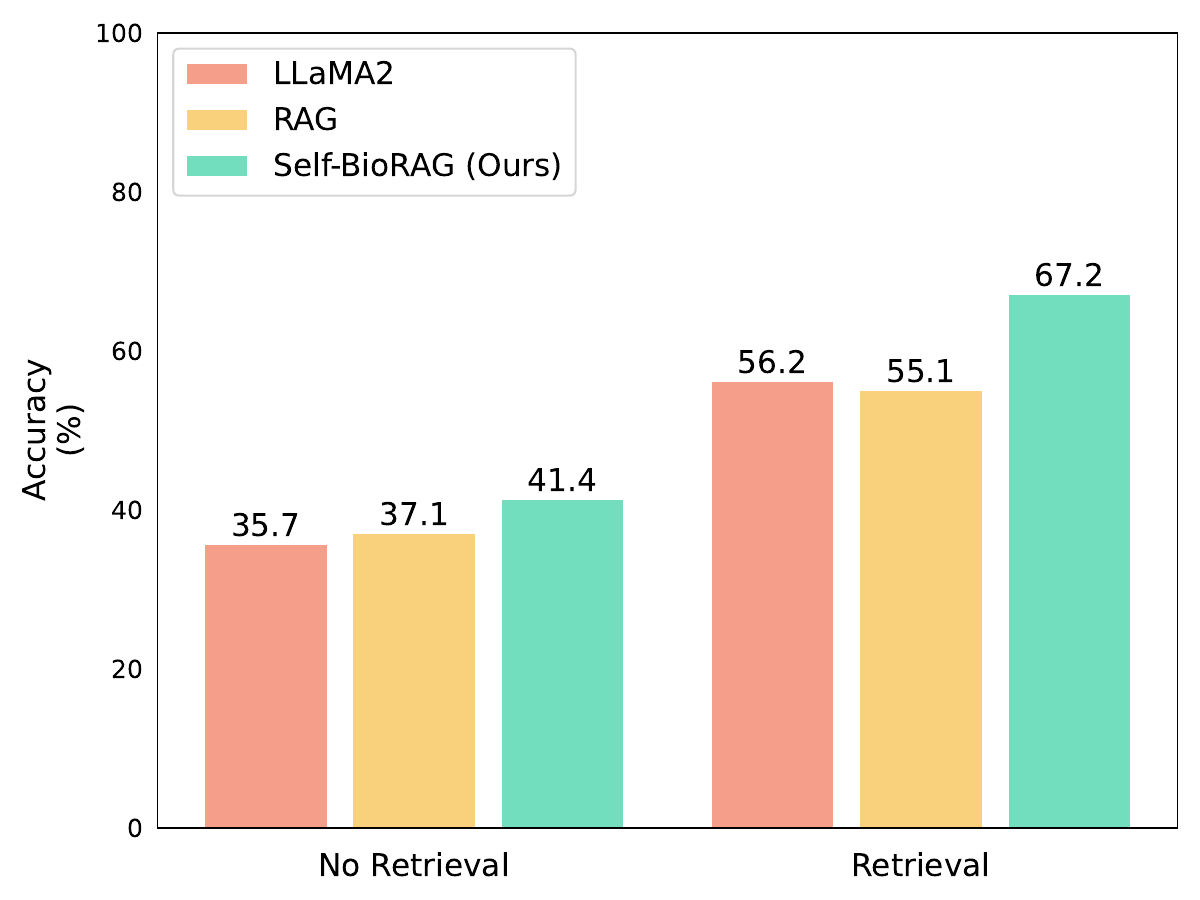}
\caption{Performance of LLaMA2, RAG, and Self-BioRAG on examples predicted as [No Retrieval] and [Retrieval] by Self-BioRAG. Self-BioRAG consistently outperforms other baselines whether or not retrieved evidence is used.}
\label{fig:selfbiorag_retrieve_noretrieve_app}
\end{figure*}

\section{Hyperparameters of Training Details}
\label{app:training_details}
We use 8 Nvidia A100 with 80GB memory to train our Self-BioRAG models.
We train our critic model for 3 epochs with a learning rate of 2e-5 with a 1\% warmup rate.
We also train our generator model for 5 epochs with a learning rate of 2e-5 with a 3\% warmup rate, and linear decay afterward.
We set our model maximum input length as 2048 for the 7B and 13B models to address at least 3 shots in training and inference.
We use Deepspeed stage 3~\citep{rajbhandari2020zero} to implement multi-GPU settings and FlassAttention~\citep{dao2022flashattention} for efficient training.
Our code is based on Self-RAG~\citep{asai2023self}, which is written in PyTorch~\citep{paszke2019pytorch} and HuggingFace~\citep{wolf2019huggingface}.
We only try once to train our models, thus it could be useful to search for better hyperparameters of our framework.

\section{How to Set Weight of Reflective Tokens}
We compute scores of each critique type as normalized probabilities of desired tokens same as Self-RAG~\citep{asai2023self}.
For the $REL$ token type, we compute as below. 
\begin{equation}
    S(REL) = \frac{p(relevant = yes)}{p(relevant = Yes) + p(relevant = No)}
\end{equation}
We compute $SUP$ using weighted scores to desirable tokens of $Fully$ $supported$ as 1 and $Partially$ $supported$ as 0.5.
We compute our $SUP$ token type as follows:
\begin{equation}
    S(SUP) = \frac{p(sup = Fully) + 0.5 * p(sup = Partial)}{p(sup = Fully) + p(sup = Partial) + p(sup = No)}
\end{equation}
We set a weighted scores $w$ of \{-1, -0.5, 0, 0.5, 1\} to the tokens of $USE=$\{1, 2, 3, 4, 5\}.
We compute $USE$ token type as follows:
\begin{equation}
    S(USE) = \sum_{i=1}^{5} w_i \frac{p(utility = i)}{\sum_{t=1}^5p(utility = t)}
\end{equation}

\section{Distinguishing When to Retrieve in MedMCQA and MMLU-Med Datasets}
\label{app:medmcqa_mmlu-med}
In Figure~\ref{fig:selfbiorag_retrieve_noretrieve_app}, we evaluate the performance of LLaMA2, RAG (LLaMA2 with MedCPT and biomedical corpora), and Self-BioRAG on examples predicted as [No Retrieval] and [Retrieval] by Self-BioRAG.
We could also observe that a similar trend is found in MedMCQA~\citep{pal2022medmcqa} and MMLU-Med~\citep{hendrycks2020measuring} datasets.
Experimental results demonstrate that Self-BioRAG consistently outperforms other baselines, whether or not retrieved evidence is used.
Interestingly, we observe that the performance of the MedMCQA dataset shows 91.1\% accuracy on retrieved questions.
This also aligns with the result that the hard constraint of using retrieved evidence in all questions shows high performance compared to not retrieving any evidence or adaptively retrieving as shown in Table 6.
We posit that the MedMCQA dataset contains questions requiring external knowledge to retrieve.

\section{Result of Closed-domain Question-answering Dataset}
We evaluate Self-BioRAG on the PubMedQA~\citep{jin2019pubmedqa} dataset, which provides a reference context to solve questions.
Our Self-BioRAG achieves 54.6\% accuracy, which is slightly lower than open foundation LMs such as LLaMA2 (56.0\%) or MEDITRON (58.1\%).
Additionally, RAG (46.2\%) and Self-RAG (52.1\%) exhibit lower performance in this trend.
We attribute this performance drop to the use of retrieved evidence during generation.
In other words, PubMedQA already provides the reference context to solve questions, and Self-BioRAG further introduces the top-10 evidence, potentially introducing errors due to excess noisy information.

\section{Detailed performance of MMLU-Med Datasets}
\label{app:mmlu-med}
To measure medical knowledge, we evaluate the Massive Multitask Language Understanding (MMLU)~\citep{hendrycks2020measuring} dataset and extract six medical and clinical domains to construct MMLU-Med benchmark as follows: Anatomy, Clinical Knowledge, College Biology, College Medicine, Medical Genetics, and Professional Medicine.
In Table~\ref{tab:mmlu_table}, we provide detailed performances of the MMLU-Med benchmark.
Our model Self-BioRAG outperforms other baselines used in the open foundation language model and open foundation model with language model on the MMLU-Med benchmark.

\begin{table*}[t]
{\resizebox{0.98\textwidth}{!}{
\begin{tabular}{ l ccccccc}
\toprule
\multicolumn{1}{ c }{} & & \multicolumn{6}{c}{\textbf{MMLU Benchmark}}     \\ \cmidrule{3-8}
 \multicolumn{1}{c}{\begin{tabular}[c]{@{}c@{}}\textbf{Model}\end{tabular}} & \multicolumn{1}{c}{\begin{tabular}[c]{@{}c@{}}\textbf{Params.}\end{tabular}}& \textbf{MMLU-AN} &
 \textbf{MMLU-CK} & \textbf{MMLU-CB} & \textbf{MMLU-CM} & \textbf{MMLU-MG} & \textbf{MMLU-PM} \\ \midrule
\textit{Open LM} \\
\midrule
Alpaca & 7B    & \multicolumn{1}{c}{44.4} & \multicolumn{1}{c}{35.5}   & \multicolumn{1}{c}{23.6}    & \multicolumn{1}{c}{31.2}      &  37.0 & 34.9  \\
FLAN-T5 & 3B     & \multicolumn{1}{c}{42.2} & \multicolumn{1}{c}{53.6}   & \multicolumn{1}{c}{41.7}    & \multicolumn{1}{c}{41.6}       &  43.0 & 43.0  \\  
PMC-LLaMA & 7B & 35.7 & 21.7 & 26.5 & 23.2 & 28.0 & 19.7 \\
Galactica & 6.7B & \multicolumn{1}{c}{40.7} & \multicolumn{1}{c}{41.5}   & \multicolumn{1}{c}{48.6}    & \multicolumn{1}{c}{34.1}       &  48.0 & 23.5  \\ 
MedAlpaca & 7B & 40.7 & 40.1 & 49.2 & 35.2 & 46.0 & 25.8 \\
MEDITRON & 7B     & \multicolumn{1}{c}{45.1} & \multicolumn{1}{c}{39.5}   & \multicolumn{1}{c}{43.2}    & \multicolumn{1}{c}{41.1}     & 38.0 & 39.8   \\
LLaMA2 & 7B     & \multicolumn{1}{c}{45.9} & \multicolumn{1}{c}{52.5}   & \multicolumn{1}{c}{38.9}    & \multicolumn{1}{c}{45.6}     & 55.0 & 40.0 \\
\midrule
\textit{Open LM + Retrieval} \\
\midrule
Self-RAG   & 7B & \multicolumn{1}{c}{42.2} & \multicolumn{1}{c}{49.1}   & \multicolumn{1}{c}{51.0}    & \multicolumn{1}{c}{39.9}     & 52.0 & 40.0   \\
Self-RAG$^{\dagger}$  & 7B & \multicolumn{1}{c}{45.2} & \multicolumn{1}{c}{55.1}   & \multicolumn{1}{c}{52.8}    & \multicolumn{1}{c}{39.9}     & 55.0 & 39.7  \\
Self-BioRAG (Ours)   & 7B & \multicolumn{1}{c}{51.1} & \multicolumn{1}{c}{60.8}   & \multicolumn{1}{c}{60.4}    & \multicolumn{1}{c}{50.9} & 52.0    &  48.2  \\ 
Self-BioRAG (Ours)   & 13B & \multicolumn{1}{c}{51.9} & \multicolumn{1}{c}{60.8}   & \multicolumn{1}{c}{66.7}    & \multicolumn{1}{c}{51.4} & 60.0    &  52.2  \\ 
 \bottomrule
\end{tabular}}}{}
\vspace{0.2cm}
\caption{Experimental results of biomedical benchmark datasets.
We use 3-shot examples to guide how to solve a benchmark example to language models.
We use the same examples described in the appendix of MedPALM~\citep{singhal2022large} for the MMLU dataset.
We refer to each dataset with the abbreviation of full name: Anatomy (AN), Clinical Knowledge (CK), College Biology (CB), College Medicine (CM), Medical Genetics (MG), and Professional Medicine (PM).
$^{\dagger}$ use biomedical corpora (e.g., PubMed, PMC, Clinical Guideline, and 
Medical Textbook) as evidence during inference.}
\label{tab:mmlu_table}
\end{table*}

\section{Prompt of Data Generation for Self-Reflection Language Model}
\label{app:critic}
We provide prompts to generate data for the self-reflection language model (critic language model).
In Table~\ref{tab:util_prompt}, utility reflective tokens are distinguished in 5 scores, ranging from 1 to 5.
We use the utility to check that the output is a useful response to the instruction.
In Table~\ref{tab:ret_prompt}, we decide if the output can be verified solely with suggested evidence on retrieval reflective tokens.
In Table~\ref{tab:rel_prompt}, we use relevant reflective tokens to determine if the evidence is relevant to the instruction and the preceding context.
Additionally, we use supportive reflective tokens to check an output sentence is supported by the evidence, described in Table~\ref{tab:sup_prompt}.

\begin{table*}[]
{\resizebox{0.98\textwidth}{!}{
\begin{tabular}{ l }
\toprule
\multicolumn{1}{ c }{\textbf{Prompt of Reflective token type Utility (USE)}} \\ \midrule
Given an instruction and an output, rate whether the response appears to be a helpful and informative answer to the query, from \\ 1 (lowest) - 5 (highest). We call this score perceived utility. \\ \\
The detailed criterion is as follows: \\
5: The response provides a complete, highly detailed, and informative response to the query, fully satisfying the information needs. \\
4: The response mostly fulfills the need in the query, while there can be some minor improvements such as discussing more detailed \\ \quad\quad information, having better structure of the response, or improving coherence. \\
3: The response is acceptable, but some major additions or improvements are needed to satisfy users' needs. \\
2: The response still addresses the main request, but it is not complete or not relevant to the query. \\
1: The response is barely on-topic or completely irrelevant. \\ \\

\#\#\# \textbf{Instruction}: Generate a description of the structure and mechanism of action of the enzyme nicotinamide adenine dinucleotide \\ phosphate (NADPH) oxidase. \\
\textbf{Output}: NADPH oxidase is a membrane-bound enzyme that plays an important role in the production of reactive oxygen species. It \\ is composed of six distinct subunits and is found in both the cytosol and the membrane. 
The subunits are composed of two copies \\each of the cytosolic subunits: p47 phox, p67 phox, p40 phox, and two copies of the membrane-associated subunits: gp91phox and \\p22phox. 
NADPH oxidase functions by donating electrons to the active site of the enzyme, thus allowing it to reduce oxygen mole\\-cules to form superoxide anion. 
The end product of this reaction is the reactive oxygen species, which is used to fight infection or \\for other body functions. NADPH oxidase is regulated by various signals such as calcium, nicotinamide adenine dinucleotide phos\\-phate (NADP), and calcium binding proteins. 
In addition, it is inhibited by certain compounds, such as apocynin, that prevent its\\ activation. NADPH oxidase plays an important role in the body's defenses against infection, and it is thus essential for many biolo\\-gical processes. \\ 
\textbf{Perceived utility}: 5 \\
\textbf{Explanation}: This response provides a descriptive explanation of the structure and function of NADPH oxidase, giving full and \\ comprehensive information. Thus, this is deemed a high utility response satisfying the user's information needs. \\ \\
\#\#\# \textbf{Instruction}: Was the enhancement of Lck activity immediate? \\
\textbf{Output}: No, the enhancement of Lck activity did not have an immediate relevance to TCR-mediated signal transduction, as judged \\from the lack of detectable de novo phosphorylation of zeta-chain and the absence of functional responses of T cells. \\
\textbf{Perceived utility}: 3 \\
\textbf{Explanation}: The response robustly addresses the aspect of Lck activity enhancement not being immediate. However, the details \\ provided thereafter are quite dense and technical, and without proper context, they may not be easy to interpret for all readers. \\Simplifying this information or providing some additional context could improve the perceived utility. \\ \\
\#\#\# \textbf{Instruction}: Alcohol intake during pregnancy causes, all, except? \\
\textbf{Output}: Brachycephaly \\
\textbf{Perceived utility}: 5 \\
\textbf{Explanation}: This is a concise and correct answer to the posed question, focusing on the exclusion of one potential effect of alcohol \\intake during pregnancy. The response specifically points out that brachycephaly is not caused by alcohol intake during pregnancy,\\ thus correctly answering the query. There's not much room for more elaborate discussion in this case, making this answer complete\\ and highly useful for the user's information needs in this context. \\ \\

\#\#\# \textbf{Instruction}: \{Example Instruction\} \\
\textbf{Output}: \{Example Output\} \\
\textbf{Perceived utility}: 
\\
\bottomrule
\end{tabular}}}{}
\vspace{0.2cm}
\caption{Prompt of Utility Reflective token. We use 3-shots to generate reflective tokens via GPT-4 API calls.}
\label{tab:util_prompt}
\end{table*}

\begin{table*}[]
{\resizebox{0.98\textwidth}{!}{
\begin{tabular}{ l }
\toprule
\multicolumn{1}{ c }{\textbf{Prompt of Reflective token type Retrieval (RET)}} \\ \midrule
You will be provided with an instruction, evidence, output sentence, and preceding sentences (optional). If the preceding sentence\\ is given, the output should be the sentence that follows those preceding sentences.
Your task is to determine whether the informat\\-ion in the output sentence can be fully verified by the evidence or if it requires further external verification. \\
If the output sentence can be verified solely with the evidence or doesn’t require any verification, respond with \textbf{[No Retrieval]}. \\
If additional information is needed to verify the output sentence, respond with \textbf{[Retrieval]}. Please provide explanations for your\\ judgments. \\ \\
\#\#\# \textbf{Instruction}: What is the mechanism of 72T4Cl activation under physiological conditions? \\
\textbf{Evidence}: Timing mechanism and effective activation energy concerned with aging and lifespan in the long-lived and thermosens\\-ory mutants of Caenorhabditis elegant.. The lifespans of many poikilothermic animals, including the nematode Caenorhabditis \\elegans, depend significantly on environmental temperature. Using long-living, thermosensory mutants of C.elegans, we tested \\whether the temperature dependency of the mean lifespan is compatible with the Arrhenius equation, which typically represents \\one of the chemical reaction rate theories. The temperature dependency of C. elegans was the Arrhenius type or normal, but daf-\\2(e1370) mutants were quite different from the others.
However, taking into account the effect of the thermal denaturation of DA\\-F-2 with the temperature, we showed that our analyzed results are compatible with previous ones. We investigated the timing me\\-chanism of one parameter (the onset of biodemographic aging (t(0))) in the lifespan equation by applying the RNAi feeding met\\-hod to daf-2 mutants in order to suppress daf-16 activity at different times during the life cycle.
In summary, we further deepened\\ the biological role of two elements, t(0) and z (the inverse of the aging rate), in the lifespan equation and mean lifespan formulated\\ by our diffusion model z(2) = 4Dt(0), where z is composed of t(0) and D (the diffusion constant).\\
\textbf{Output}: The mechanism of 72T4Cl activation under physiological conditions is not known. \\
\textbf{Rating}: [No Retrieval] \\
\textbf{Explanation}: The output sentence does not provide any factual information that needs to be verified, it is stating a fact that the mec\\-hanism is not known. The evidence does not mention 72T4Cl at all, supporting the claim that the mechanism of its activation is \\not known.
\\ \\

\#\#\# \textbf{Instruction}: Pigment providing colour to stool is \\
\textbf{Evidence}: The importance of stool tests in diagnosis and follow-up of gastrointestinal disorders in children. Stool samples should be \\evaluated macroscopically in terms of color, consistency, quantity, form, odor, and presence of mucus. The presence of a small amount\\ of mucus in stool is normal. However, the presence of copious mucus or bloody mucus is abnormal. The normal color is tawny due to \\the presence of bilirubin and bile. In infants, the stool may be green, its consistency may be watery or pasty. Stool color varies greatly \\ depending on diet. Clay-colored or putty colored stool is observed in biliary obstructions. If more than 100 mL blood is lost from the \\upper gastrointestinal system, black, tarry stool is observed. Besides bleeding, black-colored stool may also be observed due to iron or\\ bismuth treatment. Red-colored stool is observed in lower gastrointestinal tract bleeding. \\ 
\textbf{Output}: Stercobilinogen. The characteristic brown colour of feces is due to the presence of pigment stercobilin (derived from \\stercobilinogen). So, actually it is stercobilin (not stercobilinogen) which is responsible for the brown color of feces. Stercobilin is\\ derived from stercobilinogen and therefore stercobilinogen is the best answer amongst the options provided. \\
\textbf{Rating}: [Retrieval] \\
\textbf{Explanation}: The provided evidence does not mention stercobilin or stercobilinogen providing the color to stool, although it does \\mention bilirubin and bile contribute to the color. Therefore, additional evidence would be necessary to verify this output sentence. \\ \\

\#\#\# \textbf{Instruction}: \{Example Instruction\} \\
\textbf{Preceding sentences}: \{Example Preceding sentences\} \\
\textbf{Evidence}: \{Example Evidence\} \\
\textbf{Output}: \{Example Output\} \\
\textbf{Rating}: \\
\bottomrule
\end{tabular}}}{}
\vspace{0.2cm}
\caption{Prompt of Retrieval Reflective token. We use 2-shots to generate reflective tokens via GPT-4 API calls.}
\label{tab:ret_prompt}
\end{table*}

\begin{table*}[]
{\resizebox{0.98\textwidth}{!}{
\begin{tabular}{ l }
\toprule
\multicolumn{1}{ c }{\textbf{Prompt of Reflective token type Relevance (REL)}} \\ \midrule
You'll be provided with an instruction, along with evidence and possibly some preceding sentences.
When there are preceding \\ sentences, your focus should be on the sentence that comes after them.
Your job is to determine if the evidence is relevant to the \\initial instruction and the preceding context, and provides useful information to complete the task described in the instruction. \\
If the evidence meets this requirement, respond with \textbf{[Relevant]}; otherwise, generate \textbf{[Irrelevant]}. \\ \\ 

\#\#\# \textbf{Instruction}: Based on the provided genetic information, classify whether the mutation described predisposes carriers to \\breast cancer. The provided dataset is a genetic report showing a mutation in the BRCA1 gene. \\ 
\textbf{Evidence}: TBRCA mutations: is everything said? BACKGROUND: Mutations in the BRCA1 and BRCA2 genes constitute a \\risk factor for breast cancer development. BRCA mutation research has been an active field since the discovery of the genes, \\and new mutations in both genes are constantly described and classified according to several systems. AIM: We intend to provide \\an overview of the current state of BRCA1 and BRCA2 mutation description and classification. We wanted to know whether \\there was a trend towards a more frequently described mutation type and what the proportion of pathogenic mutations was.\\ RESULTS: We found that, although new mutations are described each year as reflected in current database records, very few of \\them are reported in papers. Classification systems are highly heterogeneous and a consensus among them \\
\textbf{Rating}: [Relevant] \\
\textbf{Explanation}: The evidence mentions BRCA1/2 mutation probability models and their potential role in predicting breast cancer\\ risk, directly relevant to the instruction about classifying the predisposition to breast cancer based on a mutation. \\ \\

\#\#\# \textbf{Instruction}: \{Example Instruction\} \\
\textbf{Evidence}: \{Example Evidence\} \\
\textbf{Rating}: \\

\\ \bottomrule
\end{tabular}}}{}
\vspace{0.2cm}
\caption{Prompt of Relevant Reflective token. We use 1-shot to generate reflective tokens via GPT-4 API calls.}
\label{tab:rel_prompt}
\end{table*}

\begin{table*}[]
{\resizebox{0.98\textwidth}{!}{
\begin{tabular}{ l }
\toprule
\multicolumn{1}{ c }{\textbf{Prompt of Reflective token type Supportive (SUP)}} \\ \midrule
You will receive an instruction, evidence, and output, and optional preceding sentences.  If the preceding sentence is given, the \\output should be the sentence that follows those preceding sentences. Your task is to evaluate if the output is fully supported by \\ the information provided in the evidence, and provide explanations on your judgement \\
Use the following entailment scale to generate a score: \\
\textbf{$[$Fully supported$]$} - All information in output is supported by the evidence, or extractions from the evidence. This is only applicable \\ when the output and part of the evidence are almost identical. \\
\textbf{$[$Partially supported$]$} - The output is supported by the evidence to some extent, but there is major information in the output that is \\not discussed in the evidence. For example, if an instruction asks about two concepts and the evidence only discusses either of them, \\it should be considered a $[$Partially supported$]$. \\
\textbf{$[$No support / Contradictory$]$} - The output completely ignores evidence, is unrelated to the evidence, or contradicts the evidence. \\This can also happen if the evidence is irrelevant to the instruction. \\ 
Make sure to not use any external information/knowledge to judge whether the output is true or not. Only check whether the output is\\ supported by the evidence, and not whether the output follows the instructions or not. \\ \\ 

\#\#\# \textbf{Instruction}: Ask about your symptoms and potential cardiovascular-related issues. I've been feeling intense discomfort in my \\chest during physical exertion. It sometimes shoots up to my left shoulder. There's also some mild shortness of breath. \\
\textbf{Output}: Based on the symptoms you have described, it is important to consider potential cardiovascular-related issues. Chest \\discomfort during physical exertion, shooting pain to the left shoulder, and mild shortness of breath can be indicative of a heart \\problem. These symptoms could potentially be related to angina, which is chest pain or discomfort caused by reduced blood flow \\to the heart. It commonly occurs during exercise or periods of stress. Another possibility is a myocardial infarction (heart attack),\\ where there is a blockage in one of the coronary arteries supplying blood to the heart muscle. Given the seriousness of these\\ symptoms, I would strongly advise you to seek immediate medical attention and consult with a healthcare professional for a proper\\ evaluation and diagnosis. They will be able to conduct a thorough examination, ask additional questions about your medical history,\\ possibly order tests such as an electrocardiogram (ECG) or stress test, and provide appropriate treatment recommendations based on\\ their findings. \\
\textbf{Evidence}: Disparities in Awareness of Myocardial Infarction and Stroke Symptoms and Response Among United States– and Foreign‐\\Born Adults in the National Health Interview Survey. Knowledge of MI symptoms was assessed based on responses to the question, \\“Which of the following would you say are the symptoms that someone may be having a heart attack?” Responses were pain or \\discomfort in the jaw, neck, or back; feeling weak, lightheaded, or faint; chest pain or discomfort; pain or discomfort in the arms or\\ shoulder; and shortness of breath. Similarly, knowledge of stroke symptoms was assessed by an individual’s response to the question, \\“Which of the following would you say are the symptoms that someone may be having a stroke?” Responses were sudden numbness\\ or weakness of face, arm, or leg, especially on one side; sudden confusion or trouble speaking; sudden trouble seeing in one or both \\eyes; sudden trouble walking, dizziness, or loss of balance; and sudden severe headache with no known cause. Respondents who answered \\“yes” to the 5 queries on MI and stroke symptoms were classified as knowing the symptoms of MI and stroke. Those who answered “no,” \\“don’t know,” or refused to answer were classified as not knowing the symptoms. \\
\textbf{Score}: [Fully supported] \\
\textbf{Explanation}: The output correctly identifies potential cardiovascular-related issues including angina and myocardial infarction (heart \\attack) based on the described symptoms. The evidence supports these concerns, mentioning commonly recognized symptoms of a \\heart attack including chest pain or discomfort, shooting pain in the shoulder, and shortness of breath. The output also correctly advises \\seeking immediate medical attention and consultation with a healthcare professional, which is an appropriate advice given the severity \\of these conditions. \\ \\

\#\#\# \textbf{Instruction}: \{Example Instruction\} \\
\textbf{Preceding sentences}: \{Example Preceding entences\} \\
\textbf{Output}: \{Example Output\} \\
\textbf{Evidence}: \{Example Evidence\} \\
\textbf{Score}: \\
\\ \bottomrule
\end{tabular}}}{}
\vspace{0.2cm}
\caption{Prompt of Supportive Reflective token. We use 1-shot to generate reflective tokens via GPT-4 API calls.}
\label{tab:sup_prompt}
\end{table*}

\section{Examples of Biomedical Instructions to Train Domain-Specific Instruction-Tuned Language Model}
\label{app:examples_train_generator}
We collect 84k biomedical instruction sets to train a domain-specific instruction-tuned language model.
For example, in Table~\ref{tab:generator_data}, we provide several cases used in training the language model:
(1) Our critic language model decides to not retrieve any evidence about generating an explanation of the human cardiovascular system.
(2) Retrieved evidence contains information on symptoms of Type 1 diabetes mellitus: urination, loss of weight, and stomach pain.
Therefore, our critic model predicts that the output sentence (e.g., "answer is Type 1 diabetes mellitus") fully supports the evidence.
(3) The critic model decides that more evidence is needed to answer the following question.
Even if relevant evidence is retrieved (e.g., information about cystic fibrosis), the evidence does not support solving the original question.
Overall, our critic language model predicts whether to retrieve or not based on the example instruction

\begin{table*}[]
{\resizebox{0.98\textwidth}{!}{
\begin{tabular}{ l }
\toprule
\multicolumn{1}{ c }{\textbf{Cases of Instruction used at Training Generator Model}} \\ \midrule 
\textit{No Retrieval} \\
\midrule
\textbf{Instruction}: Write a simple explanation of the human cardiovascular system suitable for a non-medical audience. <noinput> \\ \\
\textbf{Output}: [No Retrieval] It consists of the heart, blood vessels, and blood. [No Retrieval] Think of the heart as a powerful pump \\ that never stops working. [No Retrieval] Its main job is to pump blood throughout the body. [No Retrieval] The blood vessels \\ are like roads or highways that carry the blood to different parts of the body. [No Retrieval] There are three types of blood \\ vessels: arteries, veins, and capillaries. [No Retrieval] Arteries carry oxygen-rich blood away from the heart to the body's organs \\and tissues. [No Retrieval] Veins bring back oxygen-poor blood from the organs and tissues to the heart. [No Retrieval] Capillaries \\ are tiny blood vessels that connect the arteries and veins, allowing oxygen and nutrients to pass through to the cells and removes \\ waste products from them. [No Retrieval] Blood is the fuel that keeps everything running smoothly. [No Retrieval] It carries \\ oxygen and nutrients from the lungs and digestive system to all the cells in your body. [No Retrieval] It also carries waste products, \\such as carbon dioxide, away from the cells so they can be eliminated. [No Retrieval] So, the heart pumps the blood, the blood \\ vessels deliver it, and the blood carries important substances to where they're needed, keeping the body healthy. [No Retrieval] \\ In summary, the human cardiovascular system is like a transportation network that transports blood, nutrients, and oxygen to all \\ parts of our body and removes waste products. [No Retrieval] The heart acts as a pump, blood vessels are the roads, and blood is \\ the fuel that keeps everything running smoothly. [Utility:5]
\\
\midrule \midrule
\textit{Retrieval} (Fully Supported) \\
\midrule
\textbf{Instruction}: Classify whether this description corresponds to Type 1 or Type 2 diabetes mellitus. The patient showed onset of \\ symptoms during childhood, including excessive thirst, frequent urination, rapid weight loss, and stomach pain. The tests \\ confirmed low insulin levels and absence of insulin resistance \\ \\
\textbf{Output}: $[$Retrieval$]$ $<$paragraph$>$ beta cells have been destroyed. About two-thirds of body cells require insulin to absorb glucose \\ and in its absence, they will not be able to use the glucose (blood sugar) for energy. Type 1 diabetes usually comes on abruptly, \\although the damage to the beta cells may begin much earlier. Typical signs of Type 1 diabetes are a great thirst, hunger, a need to \\ urinate often, and loss of weight. To treat the disease, the person must inject insulin, follow a diet plan, exercise daily, and test blood \\ glucose several times a day. Type 1 diabetes usually occurs in children and adults who are under age 30.
This type of diabetes used \\ to be known as "insulin-dependent diabetes," "juvenile diabetes," "juvenile-onset diabetes" and "ketosis-prone diabetes." - Diabetes \\ mellitus $<$$/$paragraph$>$ $[$Relevant$]$ Based on the description provided, this corresponds to Type 1 diabetes mellitus. $[$Fully supported$]$ \\ $[$No Retrieval$]$ Type 1 diabetes mellitus is typically characterized by an early onset of symptoms, often during childhood or adolescence.\\ $[$No Retrieval$]$ Common symptoms include excessive thirst (polydipsia), frequent urination (polyuria), rapid weight loss, and \\abdominal pain.
$[$No Retrieval$]$ In addition, individuals with Type 1 diabetes typically have low insulin levels and no insulin resistance. \\ $[$No Retrieval$]$ Therefore, based on the symptoms and test results described, the answer is Type 1 diabetes mellitus. $[$Utility:5$]$
\\ \midrule \midrule
\textit{Retrieval} (No Support / Contradictory) \\
\midrule
\textbf{Instruction}: A seven year old child with recurrent chest infections and exocrine pancreatic insufficiency is suspected of having cystic \\ fibrosis. Sweat chloride levels have been observed to be between 40-60 mmol/l on two separate occasions. Which of the following test \\should be performed next to suppo the diagnosis of Cystic fibrosis? What of the following is the right choice? \\ (A) Repeat sweat chloride levels on a different day \\ (B) Demonstrate an abnormal 72 hour fecal fat collection \\ (C) Demonstrate an abnormal nasal potential difference \\ (D) Demonstrate an abnormal F508 mutation by DNA analysis\\ \\ 
\textbf{Output}: $[$Retrieval$]$ $<$paragraph$>$ to expert centers where bioassays of CFTR function like nasal potential difference measurement or \\intestinal current measurement can be done. Still, in some patients, despite symptoms compatible with CF and some indication of CFTR \\ dysfunction (e.g. only intermediate sweat chloride value), diagnostic criteria are not met (e.g. only 1 CFTR mutation identified). For these\\ subjects, the term CFTR related disorder (CFTR-RD) is used. Patients with disseminated bronchiectasis, congenital bilateral absence of \\ the vas deferens and acute or recurrent pancreatitis may fall in this category. CF has a very wide disease spectrum and increasingly the \\diagnosis is being made during adult life, mainly in subjects with milder phenotypes. In many countries, nationwide CF newborn \\ screening (NBS) has been introduced. In screen positive babies, the diagnosis of CF must be $<$$/$paragraph$>$ $[$Relevant$]$ The final answer \\is (C). $[$No support / Contradictory$]$ $[$Utility:4$]$
\\
\bottomrule
\end{tabular}}}{}
\vspace{0.2cm}
\caption{Cases of Instruction used at Training Generator Model. The Critic LM $C$ predicts retrieval of any evidence depending on the instruction. Given biomedical evidence, the model $C$ predicts whether the original output supports the evidence or not.}
\label{tab:generator_data}
\end{table*}

\end{document}